\documentclass[preprint,12pt]{elsarticle}

\usepackage{amssymb}
\usepackage{amsmath}

\usepackage[numbers]{natbib}
\usepackage{csquotes}
\usepackage{array}
\usepackage{enumerate}
\usepackage{textcomp}
\usepackage{stfloats}
\usepackage{url}
\usepackage{verbatim}
\usepackage{graphicx}
\usepackage[dvipsnames,svgnames]{xcolor}
\usepackage[export]{adjustbox} 
\usepackage{hyperref}
\usepackage{booktabs} 
\usepackage{multirow}
\usepackage{tabularx} 
\usepackage{pifont} 
\usepackage{subcaption}
\usepackage{bm}
\hypersetup{
hidelinks,
colorlinks=true,
linkcolor=blue,
citecolor=blue,
urlcolor = black
}
\usepackage{xcolor}
\usepackage{colortbl}

\def\tsc#1{\csdef{#1}{\textsc{\lowercase{#1}}\xspace}}
\tsc{WGM}
\tsc{QE}
\tsc{EP}
\tsc{PMS}
\tsc{BEC}
\tsc{DE}

\begin{document}

\begin{frontmatter}



\title{RSHallu: Dual-Mode Hallucination Evaluation for Remote-Sensing Multimodal Large Language Models with Domain-Tailored Mitigation}




\author[1]{Zihui Zhou}
\ead{zhouzihui@cqu.edu.cn}
\affiliation[1]{organization={College of Computer Science, Chongqing University},
                addressline={No. 55, South University Road, High-tech Zone}, 
                city={Chongqing},
                postcode={401331}, 
                country={China, and Heavy Rainfall Research Center of China, No. 3, Donghu East Road, Hongshan District, Wuhan, 430074, Hubei, China}}

\author[1]{Yong Feng\corref{cor1}}
\ead{fengyong@cqu.edu.cn}

\author[3]{Yanying Chen}
\ead{chenyy@cqsqxj.com}
\affiliation[3]{organization={CMA Key Open Laboratory of Transforming Climate Resources to Economy, Chongqing Institute of Meteorological Sciences},
                addressline={No. 68, Xinpaofang 1st Road, Yubei District}, 
                city={Chongqing},
                postcode={401147}, 
                country={China}}

\author[4]{Guofan Duan}
\ead{dgf@cqcst.edu.cn}

\affiliation[4]{organization={Chongqing Metropolitan College of Science and Technology},
                addressline={No. 368, Guangcai Avenue, Yongchuan District}, 
                city={Chongqing},
                postcode={402167}, 
                country={China}}
                
\author[5]{Zhenxi Song}
\ead{songzhenxi@hit.edu.cn}
\affiliation[5]{organization={School of Intelligence Science and Engineering, 
College of Artificial Intelligence, 
Harbin Institute of Technology},
                addressline={Taoyuan Subdistrict, Nanshan District}, 
                city={Shenzhen},
                postcode={518055}, 
                state={Guangdong}, 
                country={China}}

\author[1]{Mingliang Zhou}
\ead{mingliangzhou@cqu.edu.cn}

\author[6]{Weijia Jia}
\ead{jiawj@bnu.edu.cn}

\affiliation[6]{organization={BNU-UIC Institute of Artificial Intelligence and Future Networks, Beijing Normal University at Zhuhai},
                addressline={No. 18, Jinfeng Road, Tangjiawan}, 
                city={Zhuhai},
                postcode={519087}, 
                state={Guangdong}, 
                country={China}}

\cortext[cor1]{Corresponding author}

\begin{abstract}

Multimodal large language models (MLLMs) are increasingly adopted in remote sensing (RS) and have shown strong performance on tasks such as RS visual grounding (RSVG), RS visual question answering (RSVQA), and multimodal dialogue. 
However, hallucinations, which are responses inconsistent with the input RS images, severely hinder their deployment in high-stakes scenarios (e.g., emergency management and agricultural monitoring) and remain under-explored in RS.
In this work, we present \textbf{RSHallu}, a systematic study with three deliverables:
(1) we formalize RS hallucinations with an RS-oriented taxonomy and introduce \emph{image-level hallucination} to capture RS-specific inconsistencies beyond object-centric errors (e.g., modality, resolution, and scene-level semantics);
(2) we build a hallucination benchmark \textbf{RSHalluEval} (2,023 QA pairs) and enable \emph{dual-mode checking}, supporting high-precision cloud auditing and low-cost reproducible local checking via a compact checker fine-tuned on \textbf{RSHalluCheck} dataset (15,396 QA pairs);
and (3) we introduce a domain-tailored dataset \textbf{RSHalluShield} (30k QA pairs) for training-friendly mitigation and further propose training-free plug-and-play strategies, including decoding-time logit correction and RS-aware prompting.
Across representative RS-MLLMs, our mitigation improves the hallucination-free rate by up to 21.63 percentage points under a unified protocol, while maintaining competitive performance on downstream RS tasks (RSVQA/RSVG). Code and datasets will be released.
\end{abstract}




\end{frontmatter}



\section{Introduction}
\label{sec:I}


Recently, general-domain multimodal large language models (MLLMs) have rapidly advanced, catalyzing a new wave of remote sensing (RS) MLLMs \cite{GeoChat, VHM, EarthGPT, RSGPT, LHRS-Bot} for tasks such as change detection, land-use scene classification, RS visual grounding (RSVG), RS visual question answering (RSVQA), and multi-turn dialogue. These models demonstrate strong versatility and promising accuracy, suggesting a pathway toward unified RS understanding and interaction.



However, RS MLLMs also inherit a critical weakness from general-domain MLLMs: hallucinations, i.e., generated responses that are inconsistent with the given RS image. Such errors are especially problematic in RS because outputs are often consumed in high-stakes decision pipelines (e.g., emergency management and agricultural monitoring) where reliability is a first-order requirement. Beyond the general-domain setting, hallucinations in RS are further amplified by domain-specific factors: RS imagery commonly involves multi-scale objects, dense spatial relations, and diverse acquisition conditions (sensor modality and resolution), which together increase the difficulty of faithful image-grounded reasoning.


Despite extensive progress on hallucinations in general-domain MLLMs, systematic RS-oriented studies remain limited.
In particular, hallucination is not yet fully formalized as an RS-specific research target, and existing RS benchmarks that include hallucination-related evaluation are often not explicitly designed \cite{RSGPT} or to support comprehensive and efficient detection \cite{DDFAV}.
Meanwhile, several mitigation attempts rely primarily on training-based modifications that introduce additional overhead, and their mitigation effectiveness is not always directly validated under a dedicated hallucination evaluation protocol \cite{LHRS-Bot-Nova, liu2025cross}.


\begin{figure}[t]
\centering
\includegraphics[width=1\textwidth]{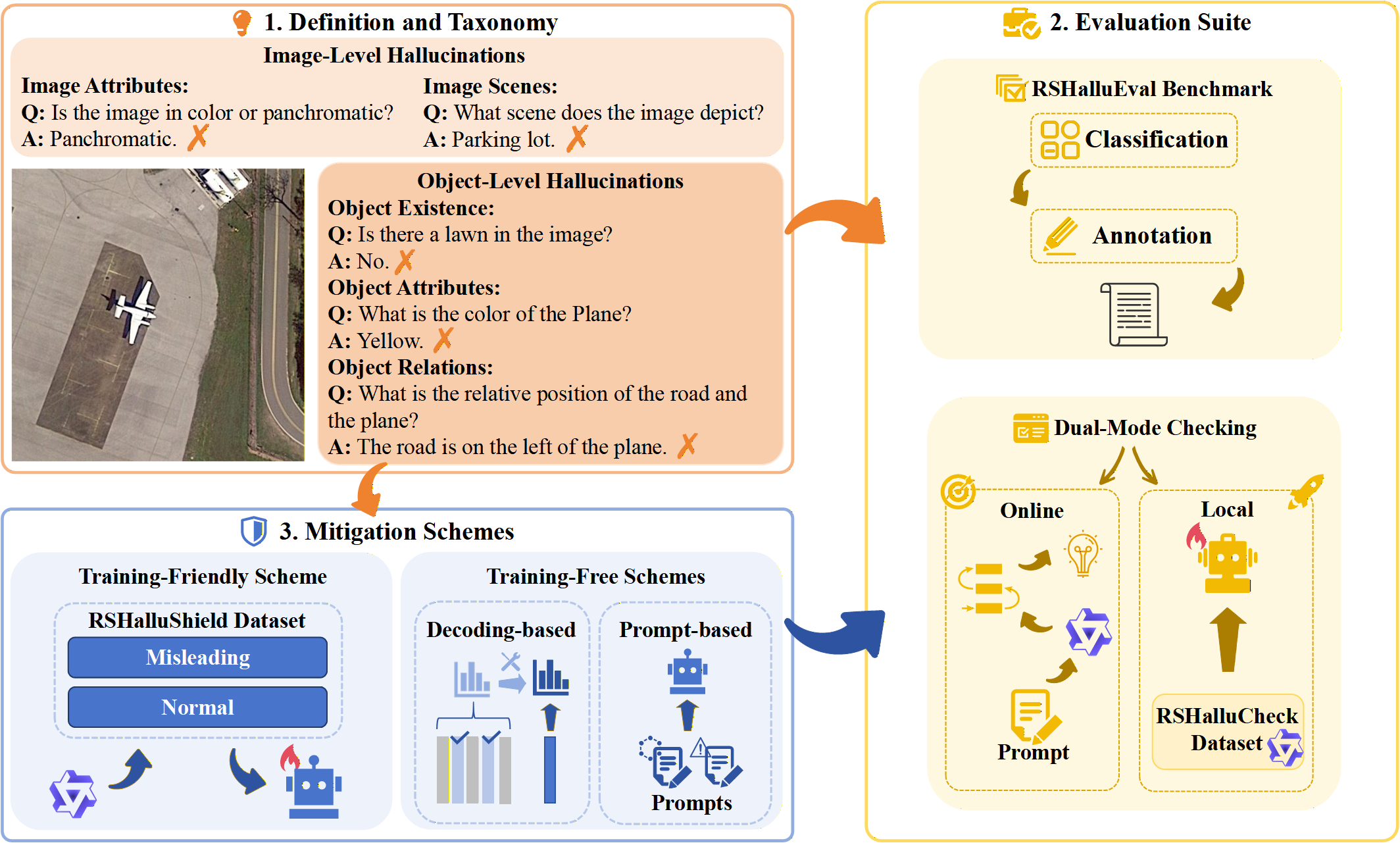}
\caption{
    Overview of our \textbf{RSHallu} pipeline for studying hallucinations in RS MLLMs. 
    We (1) establish an RS-oriented definition and taxonomy (including image-level hallucination); 
    (2) develop \textbf{RSHalluEval} with \emph{dual-mode} checking: online evaluation and a locally deployable checker fine-tuned on \textbf{RSHalluCheck}; 
    and (3) mitigate hallucinations via training-friendly fine-tuning on \textbf{RSHalluShield} and training-free plug-and-play strategies (decoding- and prompt-based).
    }
\label{framework}
\end{figure}

To bridge these gaps, we introduce \textbf{RSHallu}, a systematic study of hallucinations in RS MLLMs, and organize our investigation around three questions:\\
\textbf{Q1:} What are the characteristics of hallucinations in the RS domain, and how do they differ from those in the general domain?\\
\textbf{Q2:} How can we evaluate RS hallucinations comprehensively, accurately, and efficiently?\\
\textbf{Q3:} How can we effectively mitigate RS hallucinations in both training-friendly and training-free settings?


\noindent\textbf{Concretely,} to address \textbf{Q1}, we establish an RS-oriented definition and taxonomy of hallucinations and introduce \emph{image-level hallucination} to capture RS-specific image-relative failures beyond object-centric errors.
To address \textbf{Q2}, we develop RSHalluEval together with a \emph{dual-mode} checking protocol that supports both cloud-based auditing and locally deployable checking.
To address \textbf{Q3}, we propose RS-tailored mitigation solutions covering both training-friendly fine-tuning (via a dedicated dataset) and training-free plug-and-play strategies (via decoding- and prompt-based interventions). Fig.~\ref{framework} overviews the pipeline.

\

\begin{figure}[h]
	\centering
    \includegraphics[width=0.80\textwidth]{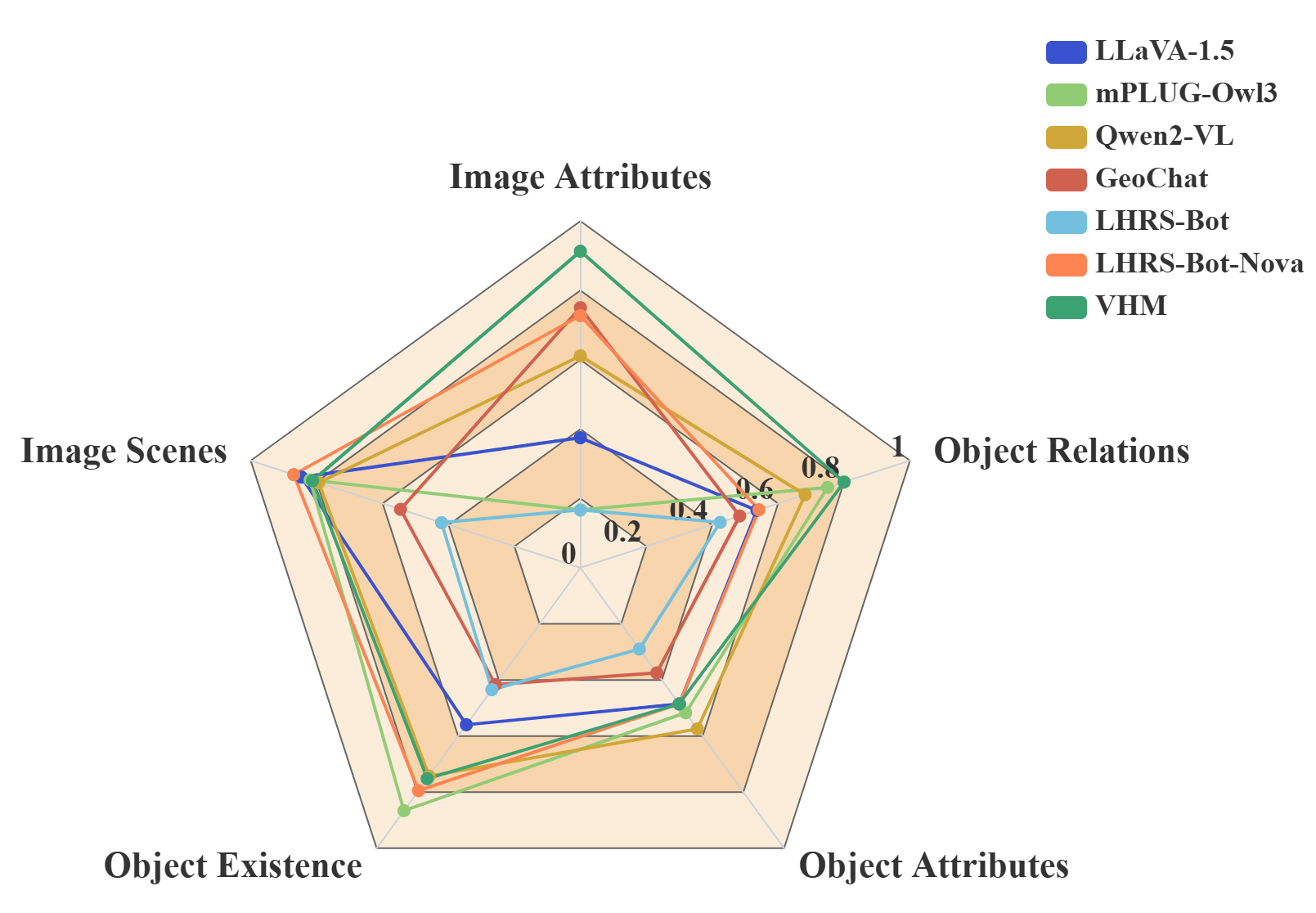}
	\caption{
    Radar plot of hallucination-free (HF) rates for representative MLLMs on \textbf{RSHalluEval}.
    HF is averaged from expert scores (higher is better) over image-level categories ($HF_{IA}$, $HF_{IS}$) and object-level categories ($HF_{OE}$, $HF_{OA}$, $HF_{OR}$), corresponding to image attributes (IA), image scenes (IS), object existence (OE), object attributes (OA), and object relations (OR).}
	\label{radar}
\end{figure}


\noindent\textbf{In particular,} our evaluation suite is guided by three principles: \emph{comprehensiveness}, \emph{accuracy}, and \emph{efficiency}.
\textbf{RSHalluEval} is manually curated from multiple RS sources, covers the defined hallucination types, and includes challenging settings such as misleading queries and negative answers.
On top of RSHalluEval, we design \emph{dual-mode} automatic \textbf{hallucination checking}:
for the \textit{\textbf{online mode}}, we perform high-precision cloud auditing via chain-of-thought prompting~\cite{CoT} with Qwen-VL-Max~\cite{Qwen-VL};
for the \textit{\textbf{local mode}}, we generate RSHalluCheck dataset by prompting Qwen-VL-Max~\cite{Qwen-VL} and fine-tune a compact checker for low cost, reproducibility, and privacy, which we use to evaluate representative RS-MLLMs on RSHalluEval (Fig.~\ref{radar}).
For \textbf{hallucination mitigation}, we cover both training-friendly and training-free regimes:
we construct \textbf{RSHalluShield} for \textit{\textbf{training-friendly}} fine-tuning, and propose plug-and-play strategies for \textit{\textbf{training-free}} settings, including decoding-time correction and RS-aware prompting.

The main contributions of this work are:
\begin{enumerate}[1)]
\item \textbf{RS-oriented hallucination formalization.} We establish an RS-specific definition and taxonomy of MLLM hallucinations and introduce image-level hallucination to capture RS image-relative failures beyond conventional object-level errors.
\item \textbf{Dual-mode evaluation suite with actionable checking strategies.} We build RSHalluEval and design flexible automatic checking strategies that support both online evaluation (for higher precision via strong MLLMs) and local deployment (for low cost, privacy, and reproducibility via a compact checker trained on RSHalluCheck).
\item \textbf{RS-tailored mitigation across training-friendly and training-free regimes.} We construct RSHalluShield for training-friendly mitigation and further propose plug-and-play mitigation strategies (decoding-based and prompt-based) that explicitly target RS-specific hallucination triggers. These methods substantially improve hallucination-free performance (up to 21.63\% HF-rate gain) while maintaining competitiveness on downstream RS tasks.
\end{enumerate}



\noindent The remainder of this paper is organized as follows: 
Section~\ref{sec:II} reviews related work; 
Section~\ref{sec:III} introduces our RS hallucination definition and taxonomy, and Section~\ref{sec:IV} presents the proposed dual-mode evaluation suite;
Sections~\ref{sec:V} and \ref{sec:VI} present training-friendly and plug-and-play mitigation schemes; Section~\ref{sec:VII} reports experiments and analysis; and 
Section~\ref{sec:VIII} concludes the paper.

\section{Related Work}
\label{sec:II}
\subsection{
Hallucinations in General-Domain MLLMs}
\label{sec:II-A}
In the field of MLLM hallucinations, research efforts have focused mainly on visual hallucinations, referring to the inconsistency between generated text and provided images. The causes of hallucinations are multifaceted, with potential causes including deficiencies in training data, visual encoders, modality alignment modules, and the inherent limitations of large language models (LLMs) \cite{liu2024survey}. Beyond these factors, current studies \cite{Opera, zoulook, Noiseboost, HaELM} widely attribute MLLM hallucination to modality bias. 

To mitigate hallucinations in MLLMs, an accurate assessment of hallucinations is important. Many studies have aimed to assess the severity of hallucinations. CHAIR \cite{CHAIR} is an early effort to assess object existence hallucinations. Li et al. \cite{POPE} proposed a benchmark and an evaluation metric that assesses hallucinations by asking about the existence of objects. Both methods focus solely on object existence hallucinations. In comparison, MMHal-Bench \cite{MMHal-Bench} covers a variety of hallucination types, including object attributes, counting, and spatial relations. In addition, Liu et al. \cite{LRV-Instruction} proposed GAVIE, which uses GPT-4 to evaluate hallucinations from two aspects: relevance and accuracy, avoiding the dependence on ground-truth labels. Other works \cite{HaELM, FaithScore, Bingo, AMBER} have also evaluated hallucinations via MLLMs.

Strategies for mitigating hallucinations in MLLMs can be categorized into data-based, model-based, training-based, and inference-based approaches \cite{bai2024hallucination}. For example, the end-of-sequence (EOS) decision \cite{EOS} was proposed to prevent training data from exceeding the perceptual limits of MLLMs by promptly terminating the process when the model reaches its perceptual threshold. Zhai et al. \cite{HallE-Switch} examined the effects of visual encoders with different resolutions on hallucinations and reported that increasing the resolution helps reduce the degree of hallucination. In addition, some methods \cite{gunjal2024detecting, sun2023aligning, yu2024rlhf} apply reinforcement learning from human feedback (RLHF) to guide the training of MLLMs to reduce hallucinations. Moreover, several studies have attempted to mitigate hallucinations by refining the decoding process \cite{Opera, zoulook, Noiseboost, wangmllm}.

\subsection{
Hallucinations in RS MLLMs}
\label{sec:II-B}
Although there have been numerous studies of hallucinations in the general field, attention to hallucinations in the RS field is only beginning. The definitive concept of hallucinations in RS is still lacking, as are comprehensive assessment methods specifically designed to evaluate hallucinations. Li et al. \cite{DDFAV} introduced the RSPOPE evaluation method based on POPE \cite{POPE} to assess hallucinations in the RS domain, but it detected only objects with hallucinations. The benchmark proposed by An et al. \cite{an2024coreval} includes a hallucination detection task, which determines the occurrence of hallucinations by checking whether MLLMs can understand objects that do not exist in the image. However, they still only considered object hallucinations. Hu et al. \cite{RSGPT} proposed the RSIEval dataset, which includes assessment metrics for the degree of hallucination; however, this dataset was not specifically designed for hallucination assessment. The data are not categorized by type of hallucinations, showing differences compared with common hallucination detection benchmarks. In contrast, the evaluation framework we propose is specifically designed for hallucination detection in RS MLLMs and more comprehensively accounts for the characteristics of hallucinations caused by the data properties and practical application features of this domain.

In terms of hallucination mitigation, Li et al. \cite{LHRS-Bot-Nova} proposed the LHRS-Instruct dataset, which relies on the design philosophy of the LRV-Instruction \cite{LRV-Instruction} dataset to mitigate hallucinations in RS MLLMs. However, they did not tailor the dataset on the basis of the distributional characteristics of RS hallucinations, nor did they directly demonstrate the effectiveness of the dataset in alleviating hallucinations. Liu et al. \cite{Co-LLaVA} proposed Co-LLaVA, which corrects LLaVA-v1.5's \cite{LLaVA} erroneous visual understanding by integrating CoCa's \cite{CoCa} visual comprehension results, thereby mitigating hallucination phenomena in remote sensing. However, their approach introduces additional structures, thereby resulting in greater time consumption. In this work, we consider the characteristics of data in the RS domain, as well as hallucinations that are prone to occur in domain transfer and practical applications, and propose training-friendly and training-free hallucination mitigation schemes.

\begin{figure}
\centering
\includegraphics[width=1\textwidth]{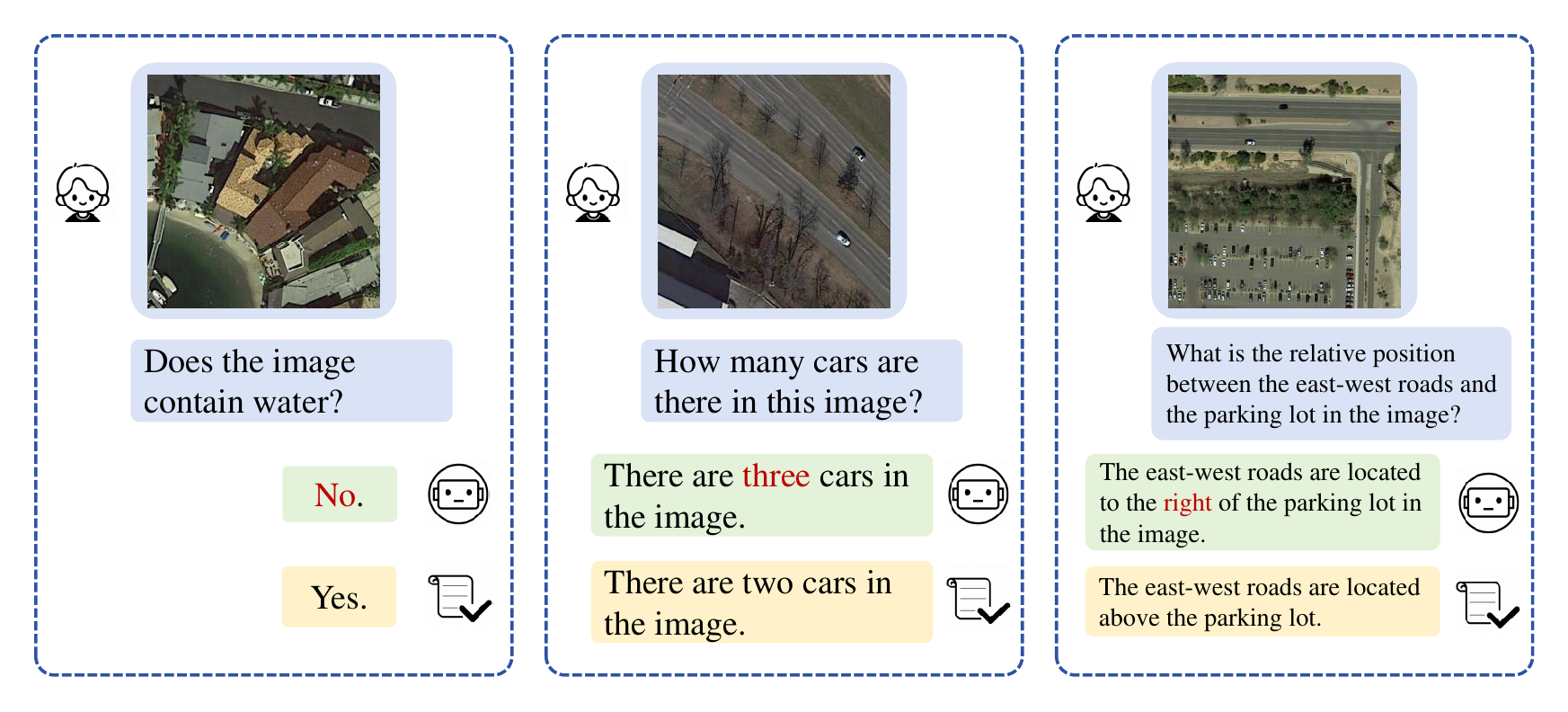}
\caption{Examples of different types of object hallucinations. From left to right are hallucinations of object existence, object attributes, and object relations. We present the hallucinatory output texts, with hallucinatory parts marked in red. The correct outputs are shown below.}
\label{other_hallucination}
\end{figure}

\section{
RS MLLM Hallucinations: Definition and Taxonomy
}
\label{sec:III}
In the general field, the focus of research on MLLM hallucinations has been on object hallucination because of the centrality of object recognition \cite{bai2024hallucination}. The MME Hallucination Subset \cite{fu2023comprehensive} is a widely used benchmark to evaluate the hallucinations of the general domain MLLMs in terms of perception and cognition, including tasks at the object level and attribute level. Other studies tend to classify object hallucination into existence, attribute, and relation hallucination \cite{gunjal2024detecting, zhai2023hallucination}.

In the RS domain, accurate identification of objects within scenes is also crucial for tasks such as object detection \cite{song2024boundary, cheng2024ef, yang2024eflnet}, change detection \cite{chen2024changemamba, li2024stade, dong2024changeclip}, RSVG \cite{zhan2023rsvg, sun2022visual}, and RSVQA \cite{wang2024rsadapter, wang2024earthvqa}. Consequently, object hallucination remains a prevalent form of hallucination in RS MLLMs, and its definition is analogous to that in the general domain. Therefore, we adopt the same classification approach as in the general domain, categorizing object hallucination into existence, attribute, and relation hallucination. Existence hallucination refers to the inaccurate detection of nonexistent or incorrect object categories in an image. Attribute hallucination is the inaccurate representation of an object's attributes, such as color, shape, and position.

Notably, relation hallucination in the RS domain is different from that in the general domain. In the general domain, relations are classified as either perceptual or cognitive \cite{Reefknot}. Perceptual relations capture spatial layouts between objects, whereas cognitive ones involve abstract verbs such as \enquote{blowing} or \enquote{watching}. In the RS domain, perceptual relations remain similar, but cognitive relations are more involved with the comparison of relative positions and attributes (such as area, length, and width) between objects. Fig. \ref{other_hallucination} depicts several types of object hallucination.

\begin{figure}[h!]
    \centering
    \includegraphics[width=0.65\textwidth]{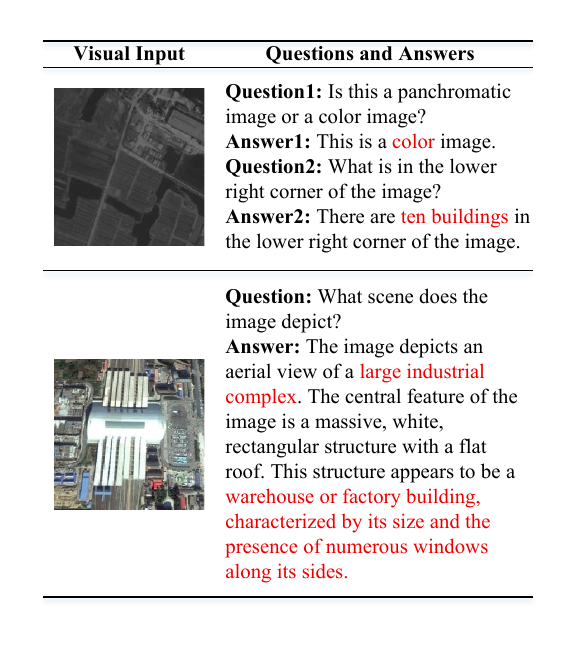}
    \caption{Some examples of hallucinations are triggered when RS MLLMs fail to comprehend image-level information. The hallucinatory parts are marked in red.}
    \label{overall_causes}
\end{figure}

Moreover, the characteristics of the RS field cause greater challenges in terms of image-level understanding. In the general field, there are a limited number of image modalities and resolutions that vary relatively little. However, image acquisition techniques in the RS field are diverse. This generates modal diversity, such as variable modalities and resolutions. Images of different modalities and resolutions have different characteristics, and the capacity to accurately discern the modality and resolution of the input image is essential for enhancing the semantic comprehension of the model. In Fig. \ref{overall_causes}, several examples of RS MLLMs for the RSVQA task illustrate this point. In the first question of the upper example, the model fails to correctly identify the modality of the image and cannot map parametric knowledge to modal semantics. Consequently, in the second question, it misjudges the features of farmland in the grayscale image as those of buildings.

Furthermore, compared with the general field, the RS field frequently encompasses tasks involving comprehensive image interpretation, such as land-use scene classification \cite{yao2023extended, roy2024cross, temenos2023interpretable} and scene understanding \cite{wang2023large, sagar2024msa}. When the model hallucinates at the overall semantic information of the RS images, it may fail to accurately identify the land use types of them as well as the spatial distribution and functional attributes of the objects (as illustrated in the lower example of Fig. \ref{overall_causes}, where the scene is misinterpreted as industrial land instead of a railway station). Consequently, its applicability will be constrained in practical RS scenarios such as high-altitude reconnaissance, agricultural resource census, and ecological environment assessment.

\begin{figure}
\centering
\includegraphics[width=3.5in]{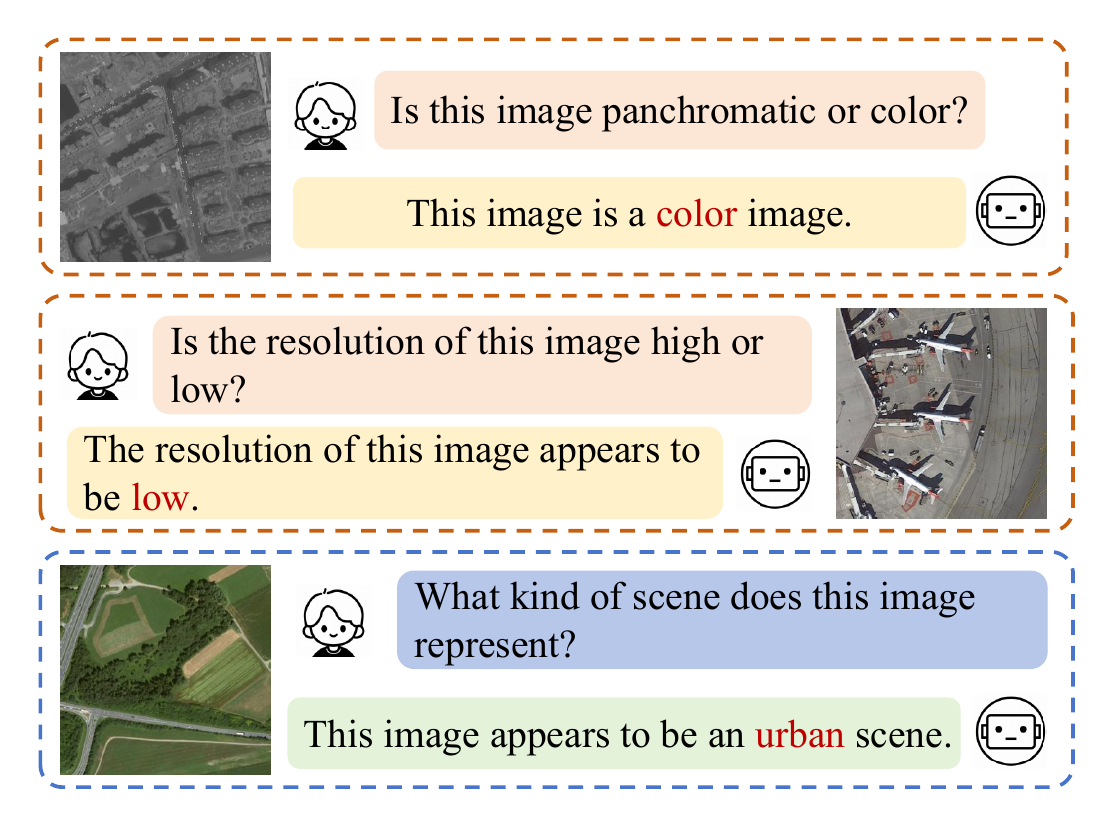}
\caption{Examples of different types of image-level hallucinations. Among them, examples with orange borders belong to image attribute hallucinations, whereas those with blue borders belong to image scene hallucinations. The hallucinatory parts are shown in red.}
\label{global_hallucination}
\end{figure}

The above two characteristics necessitate RS MLLMs to perform image-level comprehension, such as distinguishing image modalities, inferring resolution-dependent context, and understanding image scenes. On this basis, we introduce image-level hallucination, which is distinct from object-level hallucination. Image-level hallucination refers to the contradiction between the MLLMs' textual output and the input RS image in terms of understanding the overall image information. This contradiction typically involves misinterpretation of image attributes (modality, resolution, etc.) and global semantic information. We refer to the former as image attribute hallucination and the latter as image scene hallucination. Fig. \ref{global_hallucination} illustrates several examples of these hallucinations.

\section{
RSHalluEval and Dual-Mode Hallucination Checking
}
\label{sec:IV}
\subsection{
Benchmark Construction
}
On the basis of the definition of hallucinations in RS MLLMs, an RS hallucination evaluation dataset, RSHalluEval, is constructed, containing 2,023 high-quality question-answering (QA) pairs. It is built upon the RSIEval \cite{RSGPT} dataset, the RSITMD \cite{RSITMD} dataset, the UCM-Captions \cite{UCM-and-Sydney} dataset, and the Sydney-Captions dataset \cite{UCM-and-Sydney}, incorporating images from two modalities (color and panchromatic) with various resolutions and varying scenarios, including parking lots, airports, viaducts, storage tanks, harbors, etc.

While the datasets are suitable as the basis for constructing an RS hallucination detection dataset, they are not appropriate for direct employment. Previous studies \cite{RSITMD, zhou2024fine} have also shown that the sentences in these datasets may be highly similar. Moreover, in practical applications in the RS domain, factors such as lack of expertise can lead to misleading questions such as asking for the position of a ship in an image where no ship is present. MLLMs tend to have hallucinatory reactions in such challenging situations \cite{an2024coreval}. Therefore, evaluating the behavior of MLLMs when confronted with misleading questions is highly important. Although the original data include some misleading questions, the quantity is far from sufficient. Furthermore, MLLMs have been demonstrated to be more inclined to generate positive responses because of training data bias \cite{liu2024survey}. Thus, questions with negative answers are more likely to reveal the extent of hallucinations and the robustness of MLLMs.

We improve the original data in the following aspects to construct a benchmark specifically for detecting hallucinations in RS MLLMs:
\subsubsection{
Taxonomy-Aligned Categorization
}
We keep the QA pairs in the RSIEval dataset because they were manually labeled by RS experts, contain more detailed annotations, and are of higher quality. We reclassify the questions into image attributes, image scenes, object existence, object attributes, and object relations according to the types of hallucinations of the RS MLLMs.
\subsubsection{
Expert Annotation and Hard Cases
}
Three experts use the images of the four datasets to construct additional data pairs. Specifically, we label 186, 496, 100, and 298 QA pairs, respectively, via the RSIEval, RSITMD, UCM-Captions, and Sydney-Captions datasets. In the process, we introduce more misleading questions and some questions with negative answers to increase the challenge of the benchmark. Among them, the involved misleading questions including nonexistent attributes (such as color, position, number, and shape), nonexistent interactions, and nonexistent objects. In addition, for QA pairs involving resolution, we follow the evaluation criteria of RSIEval.

\begin{figure*}[!h]
\centering
\includegraphics[width=1\textwidth]{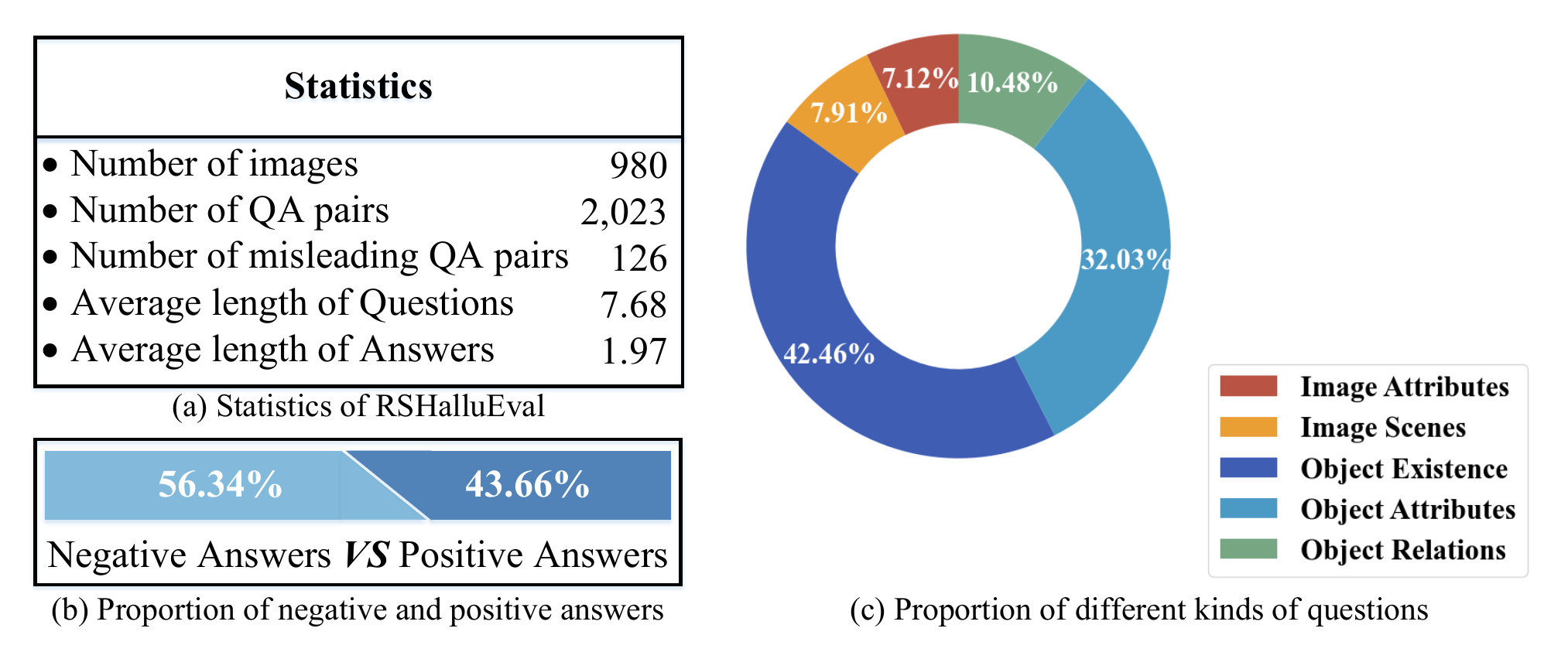}
\caption{Statistics of the proposed RSHalluEval dataset.}
\label{RSHalluEval_statistic}
\end{figure*}

Fig. \ref{RSHalluEval_statistic} presents the statistics of the proposed RSHalluEval, whereas Fig. \ref{visualize1}, Fig. \ref{visualize2} and Fig. \ref{visualize3} present several examples.

\label{sec:IV-A}
\subsection{
Dual-Mode Hallucination Checking
}
\label{sec:IV-B}
\subsubsection{
Online Mode: API-based Auditing with CoT
}
To evaluate the extent of hallucinations in RS MLLMs, a natural approach is to leverage zero-shot evaluation via a large-scale MLLM. We select Qwen-VL-Max, which is the best-performing model in the Qwen-VL series. Compared with the open-source version of Qwen-VL, it achieves comparable performance to Gemini Ultra \cite{Gemini} and GPT-4V \cite{GPT-4V} across multiple multimodal tasks, significantly outperforming the best results of prior open-source MLLMs.

Although Qwen-VL-Max has strong capabilities, it is designed primarily for general-domain applications. To facilitate its adaptation to RS scenarios, we explicitly restrict the application context to the RS domain during prompting. Other inputs include the RS image, the question, the answer output by the MLLMs, and the ground truth answer.

Despite this, Qwen-VL-Max still fails to detect hallucinations effectively in some scenarios. We attribute this partly to the fact that determining whether hallucinations have occurred requires comparing the image, the generated text, and the ground truth, which requires an inference process. Therefore, simply allowing Qwen-VL-Max to output responses directly may lead to inaccurate judgments. Moreover, this is also due to the insufficient adaptation of Qwen-VL-Max to the RS domain. We consider improving the evaluation scheme from the former aspect, guiding CoT to ask the model to reason explicitly step by step. Specifically, we add the instruction \enquote{Let us think step by step.} to the prompt.
We compare the changes in evaluation accuracy before and after the improvement in Table \ref{Etable2}, which further demonstrates that our improvements, despite their simplicity, yield enhancements.


\subsubsection{
Local Mode: Compact Checker Fine-tuned on RSHalluCheck
}
Although online API invocation for detecting hallucinations in RS MLLMs has shown high accuracy, this method incurs network latency and data transmission costs because of the necessity of data transmission over the network. Moreover, as the most powerful version of the Qwen series, Qwen-VL-Max may have a large number of parameters; thus, its processing speed is slower than that of smaller models. Therefore, we aim to train a more compact checker specifically designed to evaluate hallucinations in the RS domain. Unlike the online mode, we do not provide ground truth answers to accommodate a broader range of application scenarios.

To train the checker, we propose an RS hallucination detection dataset, RSHalluCheck. We construct RSHalluCheck on the basis of the RSICap dataset \cite{RSGPT} because the images are manually annotated by five RS experts and are of higher quality. More importantly, unlike other RS image–text datasets, the RSICap dataset has more fine-grained captions, with an average of 60 words per caption. The detailed captions facilitate the creation of diverse QA pairs.

To obtain a dataset of considerable scale, we generate QA pairs via automated annotation. The initial data generation approach provides the RS image and caption, prompting Qwen-VL-Max to generate QA pairs of specified types and quantities. Concurrently, we pose the following question: MLLMs exhibit insufficient context attention \cite{wang2024vigc, lee2023volcano}; that is, when generating extensive text, they tend to prioritize the text adjacent to the generation position while neglecting the preceding image features, thereby yielding generated content that is incongruent with the image information. Alternatively, given that the caption provides a highly detailed description of the image, we inquire whether it is possible to generate QA pairs using solely the caption. The latter approach can mitigate certain hallucination issues and, by not involving the vision modality, may achieve better generation efficiency. Therefore, we attempt to prompt Qwen-VL-Max to generate QA pairs using only fine-grained captions to generate questions about image attributes, image scenes, object existence, and object attributes. For questions concerning object relations, given the infrequent involvement of relative positions and comparisons between objects in the captions, we utilize both the image and the fine-grained caption to prompt Qwen-VL-Max to generate QA pairs. We extract 100 QA pairs from the generated data for validation, achieving an accuracy rate of 92\%, which substantiates the rationality of our generation approach. We assume that six QA pairs are generated each time, with the first three being QA pairs without hallucinations and the last three exhibiting hallucinations in their answers.



Owing to the language style of Qwen-VL-Max, the generated answers tend to be extensive and incorporate explanations for the answers. However, given the simplicity of answers in existing RS multimodal datasets, most responses from fine-tuned RS MLLMs are relatively concise. To enable the trained RS hallucination checker to accurately assess hallucinations from MLLMs of different styles, we modify some long QA pairs that begin with \enquote{Yes/No} to simply \enquote{Yes} or \enquote{No}, enhancing the diversity of language styles in RSHalluCheck to some extent.

\begin{figure}
\centering
\includegraphics[width=1\textwidth]{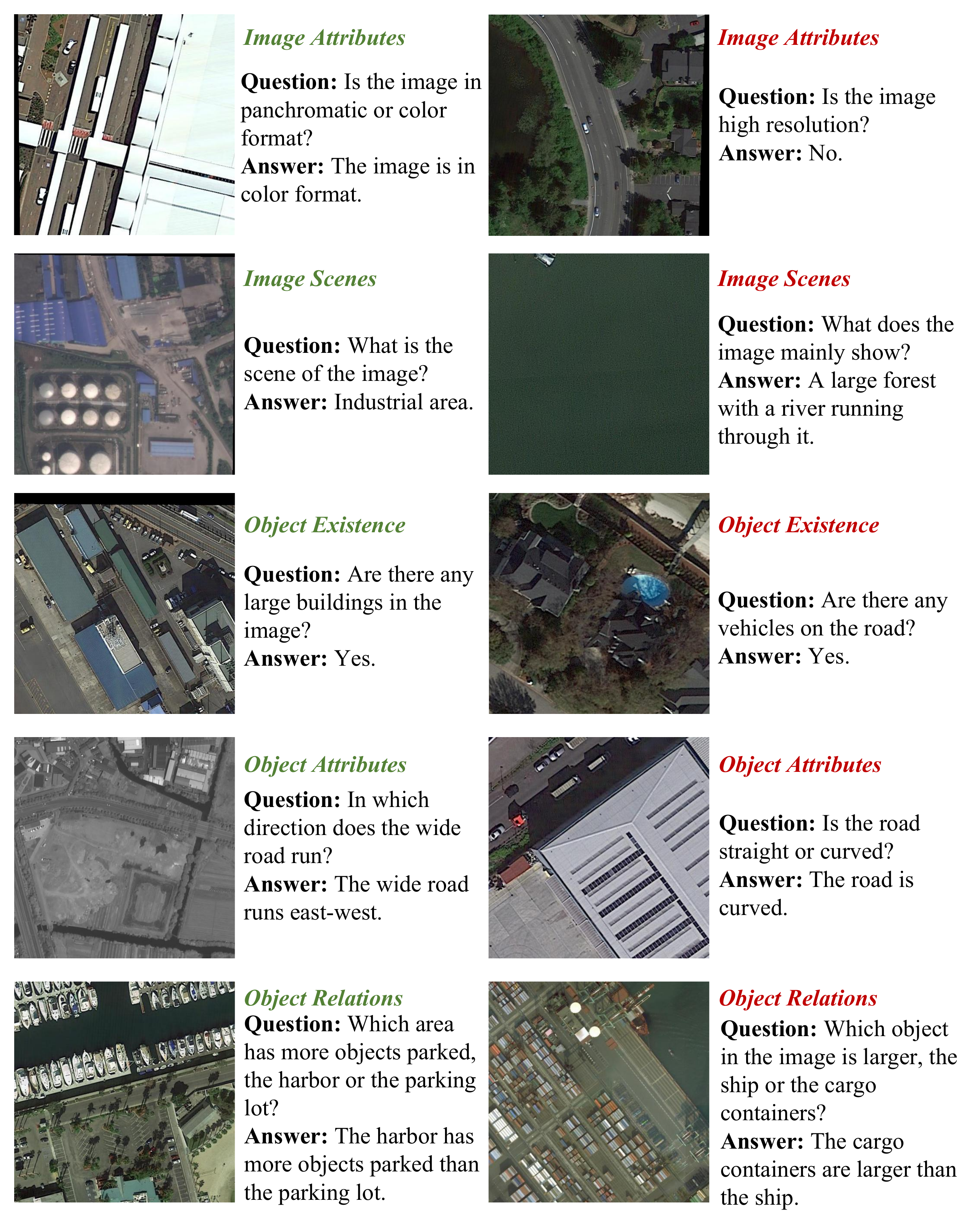}
\caption{Examples of QA pairs in the RSHalluCheck dataset. Red font indicates examples with hallucinatory answers, whereas green font indicates examples with nonhallucinatory answers.}
\label{checker_example}
\end{figure}

We randomly select 1,261 images from RSICap for data generation, resulting in 15,396 QA pairs. These data are divided into a training set and a validation set. The training set contains 1,000 images and 13,994 QA pairs, with 84, 2,079, 6,407, 4,779, and 645 QA pairs for image attributes, image scenes, object existence, object attributes, and object relations, respectively. The validation set contains 261 images and 1,402 QA pairs, with the numbers of QA pairs for image attributes, image scenes, object existence, object attributes, and object relations being 14, 190, 600, 450, and 148, respectively. To prevent data leakage, all QA pairs from the same image are placed in the same subset. Several examples are presented in Fig. \ref{checker_example}.

\section{
Training-Friendly Mitigation via RSHalluShield
}
\label{sec:V}
Data-based hallucination reduction methods \cite{gunjal2024detecting, LRV-Instruction, xiao2025detecting} alleviate hallucination by proposing datasets to fine-tune MLLMs, which can be easily adapted to different MLLMs. Therefore, we adopt this kind of scheme, proposing the RSHalluShield dataset to mitigate the hallucination phenomenon in RS MLLMs. Similar to the RSHalluCheck dataset, we select 1,292 images from the RSICap dataset for construction, none of which are used in RSHalluCheck. We construct this hallucination mitigation dataset, which is tailored to the characteristics of RS MLLMs, via the following steps.
\subsection{
Normal vs. Misleading Ratio
}
\label{sec:V-A}
Li et al. \cite{POPE} found that MLLMs tend to answer \enquote{yes} directly when responding to questions. Liu et al. \cite{LRV-Instruction} pointed out that this phenomenon arises because most models are fine-tuned on imbalanced datasets containing only positive instructions, leading to biases in the inference process. On this basis, they proposed the LRV-Instruction dataset, which includes an equal number of positive and negative examples. Motivated by their approach, we incorporate both types of QA pairs into the RSHalluShield dataset.

In the LRV-Instruction dataset, positive examples refer to QA pairs where the questions are consistent with the image semantics, whereas negative examples involve misleading instructions with nonexistent elements, existing objects with wrong attributes, and incorrect knowledge, with answers that highlight unreasonable aspects. To make this more intuitive, we refer to these two types of QA pairs as normal and misleading examples, respectively, in the following sections. We conduct an experiment (Section~\ref{sec:VII-G}) to determine the ratio to be 1:1 to better adapt this approach to the RS domain.

\subsection{Category-Specific Data Generation}
\label{sec:V-B}
For normal examples, we subdivide them into image attributes, image scenes, object existence, object attributes, and object relations. For each type of question, we employ corresponding prompts to enable Qwen-VL-Max to generate QA pairs of specific types and quantities on the basis of RS images and fine-grained captions. This process is similar to that of RSHalluCheck.

For misleading examples, we prompt the MLLM to generate QA pairs on the basis of the following principles:

\begin{itemize}
\item The misleading information in the questions encompasses nonexistent attributes (such as color, position, number, and shape), nonexistent interactions, and nonexistent objects.
\item The objects referenced in the questions should be those commonly encountered in the RS domain. Accordingly, we provide the MLLM with a list of common objects.
\item The objects referenced in the questions should be relevant to the RS images; that is, it is reasonable for them to appear in the RS images.
\item The language style must be diversified to ensure the universality of the data.
\item The answers must be accurate, not only to avoid being misled but also to correct the issues in the questions.
\end{itemize}

\begin{figure}[h!]
\centering
\includegraphics[width=1\textwidth]{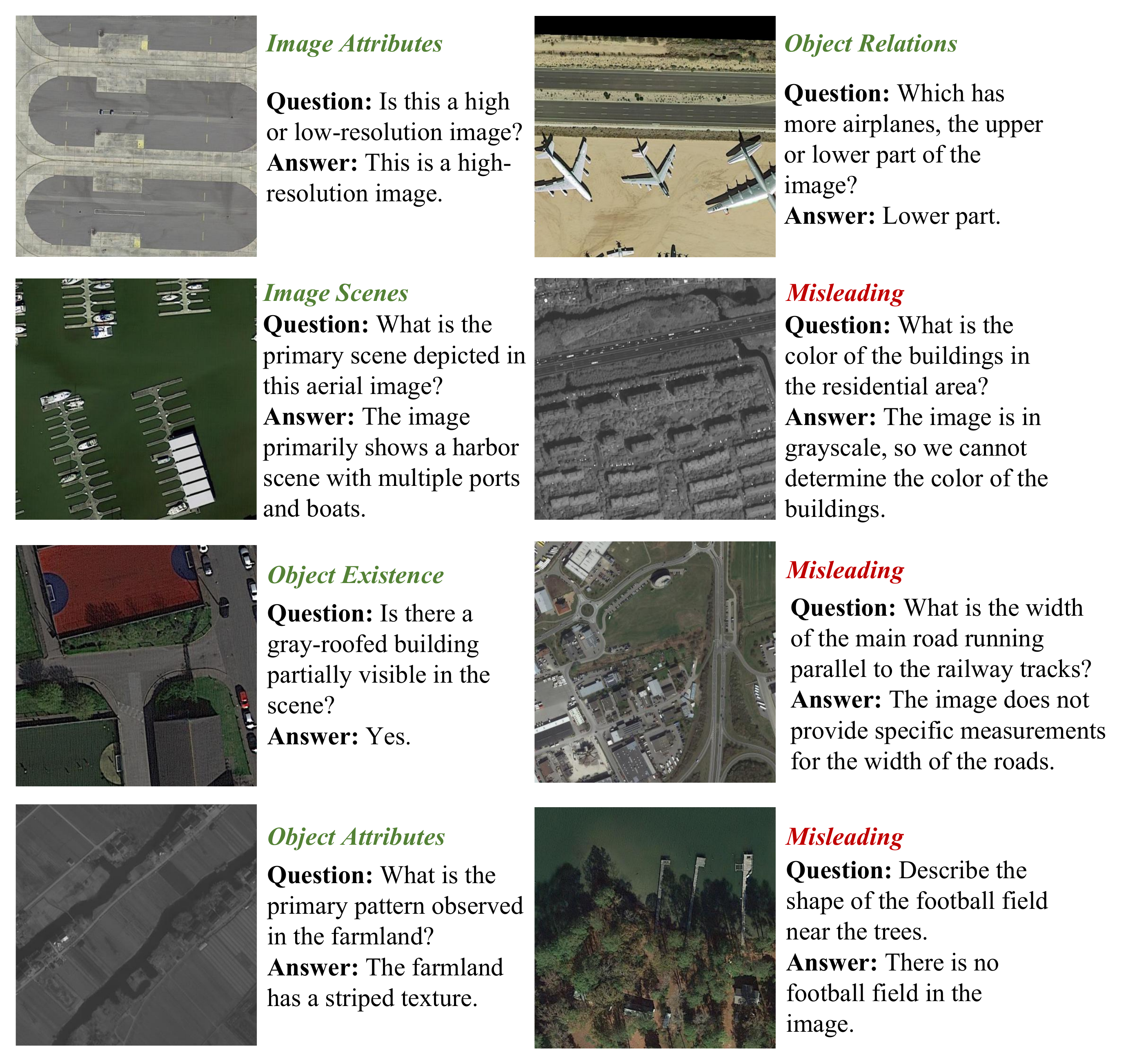}
\caption{Examples of QA pairs in the RSHalluShield dataset. Red font indicates normal examples, whereas green font indicates misleading examples.}
\label{shield_example}
\end{figure}

\subsection{Quality Control}
\label{sec:V-C}
We filter out the data and remove those with long answers. Additionally, we find that some QA pairs mention the caption we provide to the MLLM, which is not available in actual model training. Therefore, we paraphrase such QA pairs to ensure that they do not reference the caption and maintain semantic correctness. After that, we randomly examine 100 QA pairs, and the proportion of correct answers reaches 86\%.

After the aforementioned processing, we obtain a dataset comprising 30,000 QA pairs, with 15,000 normal and 15,000 misleading examples each. Some examples of the dataset are shown in Fig. \ref{shield_example}.

\section{
Training-Free Mitigation: Plug-and-Play Strategies
}
\label{sec:VI}
Although training-based methods can effectively reduce the hallucination of MLLMs, they are limited in scenarios where source code cannot be obtained or computational resources are scarce. Therefore, we consider exploring approaches to alleviate hallucinations that do not require training.

\subsection{
Decoding-time Logit Correction
}
\label{sec:VI-A}
To mitigate hallucinations without additional training, several studies have alleviated hallucinations through contrastive decoding \cite{Convis, leng2024mitigating, wang2024mitigating}. However, most contrastive decoding methods require performing multiple inferences on a single sample to obtain contrastive signals, resulting in a significant reduction in inference efficiency. In contrast, Wang et al. \cite{wangmllm} observed that the ability of MLLMs to recognize objects in shallow decoders deteriorates with depth, resulting in hallucinations. They therefore proposed a shallow, knowledge-guided, dynamic calibration decoding strategy that corrects final-layer logits using shallow-layer knowledge during autoregressive generation.

\begin{figure}
\centering
\includegraphics[width=0.95\textwidth]{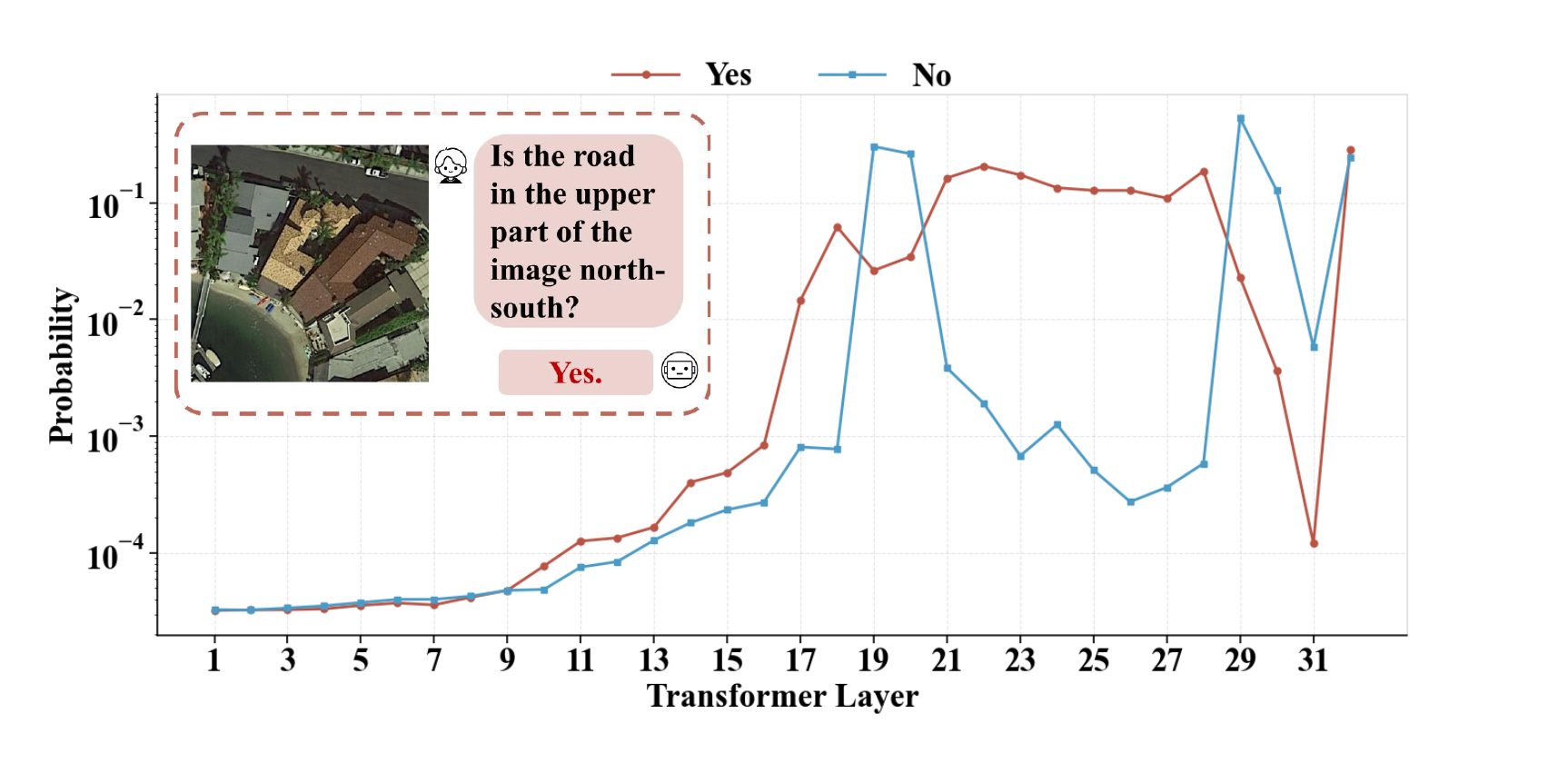}
\caption{An example of the probability distributions of the ground truth token \enquote{No} and the hallucinatory token \enquote{Yes} across all layers of an MLLM when it generates a hallucination answer in the RS domain. Red text indicates the hallucinatory response.}
\label{probs_across_layers}
\end{figure}

Motivated by Wang et al. \cite{wangmllm}, we investigate whether MLLMs exhibit similar behavior in the RS domain. Fig. \ref{probs_across_layers} shows that when generating RS answers, the correct token \enquote{No} attains high probability in middle layers (e.g., 19, 20, and 29-31), yet is overtaken by \enquote{Yes} at the final layer, resulting in hallucination. Moreover, while Wang et al. \cite{wangmllm} favor shallower features as references, this example demonstrates the potential of leveraging some deeper layers for hallucination correction.

Based on this finding, we aim to design a logit correction strategy that identifies more reliable intermediate layers to rectify output logits. Wang et al. \cite{wangmllm} argued that MLLM hallucinations arise because deep layers over-allocate attention to the textual modality, generating outputs that deviate from visual information. Yin et al. \cite{Clearsight} investigated the attention allocation patterns of each model layer with respect to three input components: system instructions, visual features, and user instructions. They found that the model focuses far less on the latter two, especially the visual modality. Meanwhile, the final layer shows an increased attention to visual and user instructions. We argue this shift may alter the model’s focus and thus change the logit distribution. Therefore, we propose the following metric based on attention weights for visual and user instructions:
\begin{equation}
    a_b^i = \frac{a_v^i}{a_t^i} - r_p \times a_t^i
    \label{decoding_metric}
\end{equation}
where $i$ denotes the index of the decoding layer, $a_v^i$ and $a_t^i$ represent the attention weights allocated to visual information and user instructions, respectively. $r_p$ is the penalty factor that penalizes the attention weights allocated to user instructions. $a_{b}^i$ reflects the attention allocation of the current decoding layer to visual features and user instructions.

We select a set of intermediate layers $\{m_i,m_{i+1},...,m_n\}$ as the initial reference layers $M_{origin}$, and compute $a_b^i$ for these layers as well as the final layer $a_b^{last}$. We then identify the top $K_m$ reference layers with $a_b^i > a_b^{last}$, which are deemed more dependent on visual information than the final layer.

After that, an intuitive correction strategy is to weight the logits of these selected layers against those of the final layer. However, intermediate layers tend to lack full semantic modeling capabilities and are not specifically optimized for stop signals during training. This may lead to substantial discrepancies between their predicted logit distributions and that of the final layer, as well as failure to generate semantic stop signals. Therefore, a second round of screening is required prior to correction. Specifically, we extract the top $K_t$ highest-probability tokens from the outputs of each candidate layer as well as the final layer. We then conduct further filtering on these tokens. A candidate layer is retained if any token among its top $K_t$ tokens has a probability $p$ in the final layer exceeding the threshold $thred_t$. The final layers yielded from this filtering process are the layers for correction, denoted as $M_{final}$.

Subsequently, we perform a softmax operation on $a_b^i$ of each layer in $M_{final}$ to obtain the weight $W_f$ for each layer, conducting weighted summation on the logits to derive the logits for correction:
\begin{equation}
    logit_{ref} = \sum_{i=1}^{|M_{final}|} logit_{i} \times w_{f}^i
\end{equation}
where $logit_{i}$ denotes the logit of the $i$-th candidate layer in $M_{final}$, $w_{f}^i \in W_f$ is its corresponding weight, and $logit_{ref}$ represents the logit for correction.

\begin{figure}
\centering
\includegraphics[width=1\textwidth]{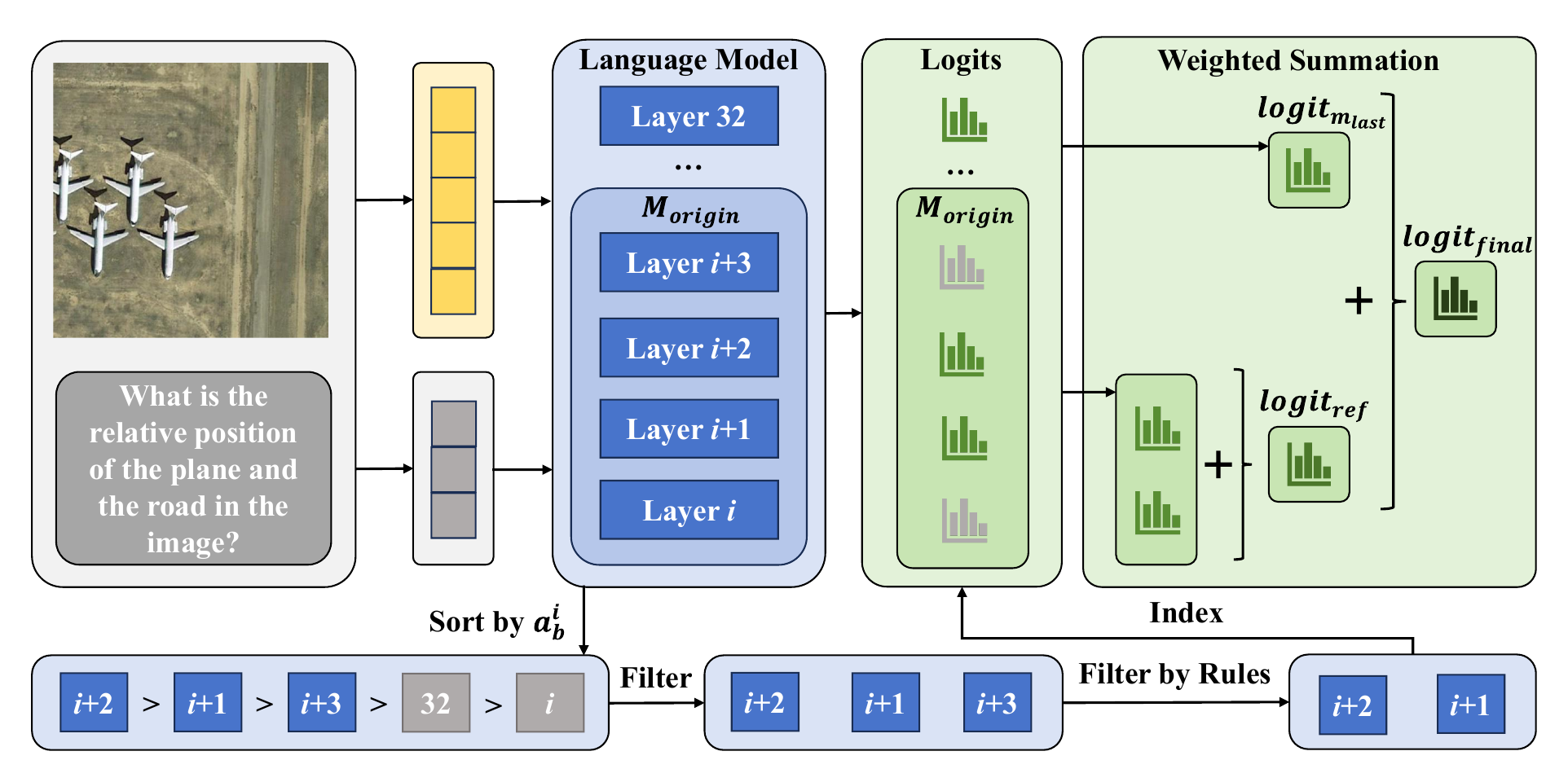}
\caption{Illustration of the proposed decoding-based scheme.}
\label{decoding_method}
\end{figure}

Finally, a weighted summation is applied to the logit of the final layer with the recall rate $r$:
\begin{equation}
    logit_{final} = (1 - r) \times logit_{m_{last}} + r \times logit_{ref}
\end{equation}
Where $logit_{m_{last}}$ denotes the logit of the final layer and $logit_{final}$ is the corrected logit. This method is illustrated in Fig. \ref{decoding_method}.

\subsection{
RS-Aware Prompting
}
\label{sec:VI-B}
In addition to decoding-based methods, we propose RS-aware prompting strategies that target two practical causes of hallucinations: insufficient global perception and vulnerability to counterfactual (misleading) queries.

\subsubsection{Overall Perception Prompting}
\label{sec:VI-B1}



One important reason for hallucinations in MLLMs is insufficient ability to extract visual features and biased attention between modalities, especially in RS scenes with multiscale objects.
To enhance global perception and encourage vision-grounded reasoning, we design an overall understanding prompt $p_g$:
\textit{Describe the picture concisely in one sentence, including the type of land use and the main targets and their locations.}
Given an image-question pair $\{v,t\}$, we first query the model with $\{v,p_g\}$ to obtain a concise global description $o_g$.
We then concatenate $o_g$ with the original question $t$, and feed them together with $v$ into the MLLM to produce the final answer.
This two-stage prompting injects vision-grounded context into the textual input and helps balance cross-modal attention.

\subsubsection{
Counterfactual Perception Prompting
}
\label{sec:VI-B2}



MLLMs are also prone to misinterpreting counterfactual inputs, which are common in RS applications due to users' limited domain expertise (e.g., querying nonexistent objects or incorrect attributes).
To make the model more cautious under such conditions, we prepend the following instruction to the input:
\textit{You are a remote sensing expert. Please answer the following question. Note that the question may contain counterfactual information, such as objects that do not exist in the image or attributes that are inconsistent with the image: $t$.}
This strategy explicitly alerts the model to verify the premise of the question against the image before answering.

\begin{figure}[h]
\centering
\includegraphics[width=1\textwidth]{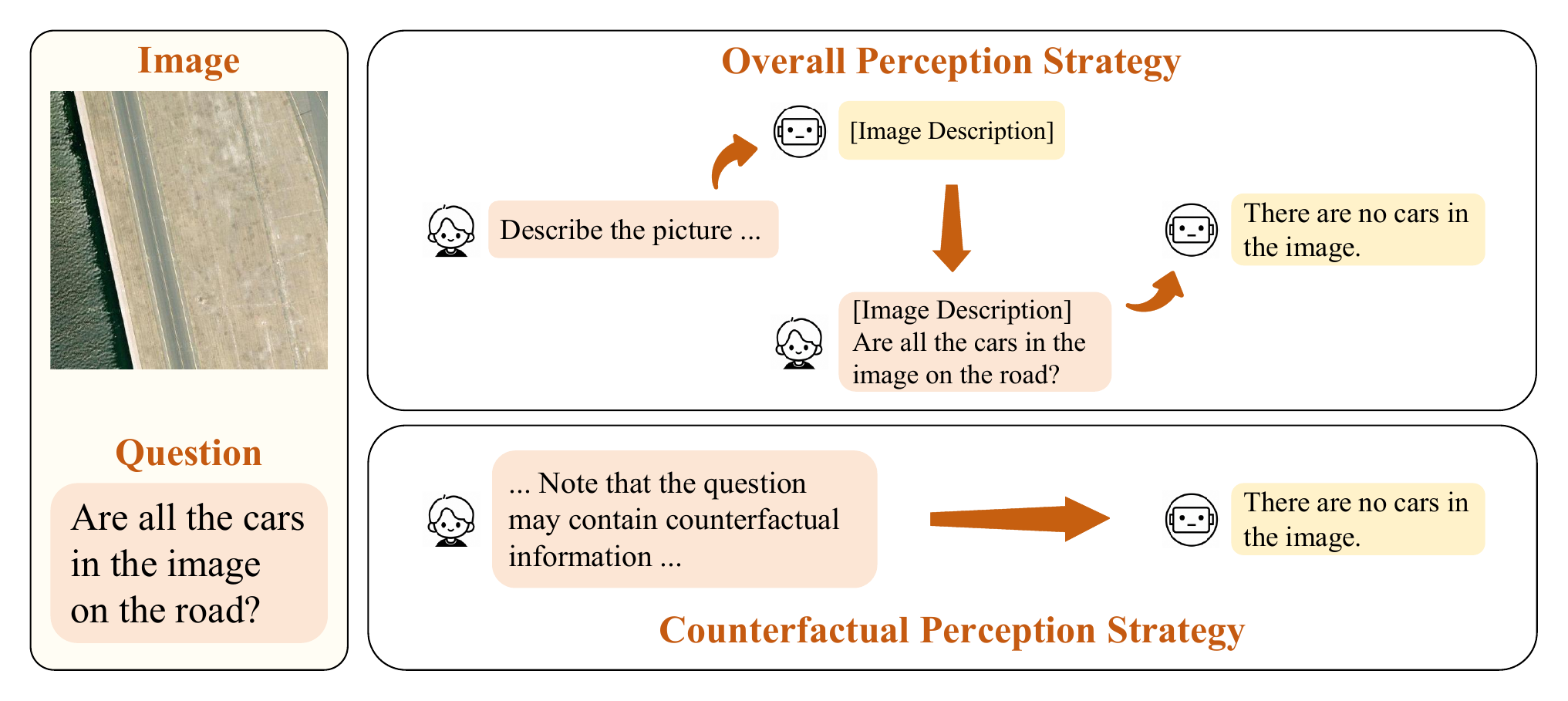}
\caption{Illustration of the proposed prompt-based schemes.}
\label{prompt_method}
\end{figure}



For convenience, we refer to the above two strategies as the \enquote{overall perception strategy} and the \enquote{counterfactual perception strategy}. Fig.~\ref{prompt_method} illustrates their inference procedures.
Unlike training-based mitigation, these prompting strategies are plug-and-play and require no additional supervision or model updates, making them readily applicable to black-box RS MLLMs.
While several training-free hallucination mitigation methods have been explored in the general domain~\cite{qu2024look, yin2024woodpecker}, our prompts are tailored to RS-specific failure modes (e.g., global land-use understanding and counterfactual/misleading RS queries), and are designed to improve robustness under practical RS deployment conditions.

\section{Experiments and Discussion}
\label{sec:VII}
\subsection{Experimental Setup}
\label{sec:VII-B}
For the evaluation of MLLM hallucinations in the RS, we use expert assessment and Qwen-VL-Max's API, which is based on CoT, and Qwen2-VL \cite{Qwen-VL}, which is fine-tuned with RSHalluCheck. In the expert evaluation, we randomly select 300 QA pairs from each model for assessment. On the other hand, automated evaluation assesses the answers of each model to all the questions. We employ all the evaluation strategies mentioned above in Table \ref{Etable1}, whereas in the remaining cases, we utilize the CoT-activated automated evaluation strategy to ensure accuracy and evaluation efficiency.

For the hallucination checker, we initially train three models: LLaVA-1.5 \cite{LLaVA}, Qwen2-VL, and mPLUG-Owl3. We select the fine-tuning Qwen2-VL as the hallucination checker. We utilize the AdamW optimizer for optimization, with $\beta_1$, $\beta_2$, and $\epsilon$ set to 0.9, 0.999, and $10^{-8}$, respectively. The learning rate is adjusted via a cosine annealing schedule, with an initial learning rate of 0.0001 and no warmup steps. We train the models for 2 epochs on the RSHalluCheck dataset with a batch size of 2, and we employ LoRA during training, setting the rank to 8 and the alpha to 32.

For the decoding-based hallucination mitigation strategy, we set $r_p=0.1$, $K_m=2$, and $K_t=2$. When applying this strategy to LLaVA-1.5, GeoChat, and VHM, the threshold $p$ and retracing ratio $r$ vary across these MLLMs. For LLaVA-1.5, $p$ is set to 0.3 and 0.05, while $r$ is configured as 0.8 and 0.5, corresponding to greedy and sample decoding, respectively. For GeoChat, $p = 0.2$ and $r = 0.3$ for both greedy and sample decoding, with consistent values across the two modes. For VHM, $p$ takes values of 0.05 and 0.2 for greedy and sample decoding, respectively, whereas $r$ is fixed at 0.7 for both decoding modes. Across all these MLLMs, $M_{origin}$ are configured as $\{29, 30, 31\}$.

We conduct LoRA fine-tuning of Qwen2-VL on RSHalluShield to mitigate its hallucination level, with the rank set to 8 and alpha set to 32. Optimization is performed via the AdamW optimizer, where $\beta_1$, $\beta_2$, and $\epsilon$ are set as 0.9, 0.999, and $10^{-8}$, respectively. The learning rate is adjusted via a cosine annealing schedule, initialized at 0.0001 with no warmup steps. The model is trained for 3 epochs with a batch size of 2.

Unless otherwise specified, the number of parameters for the locally deployed MLLMs used in this study is 7B. Additionally, we invoke the Qwen-VL-Max API, for which the number of parameters remains undisclosed.

\subsection{Metrics}
\label{sec:VII-A}
We use the Hallucination-Free (HF) rate to evaluate the extent of hallucination mitigation in MLLMs on the basis of the scores assigned to each QA pair. For human evaluation, we allow RS experts to score answers as 0, 0.5, or 1 point, with a score of 0 indicating that the answer has obvious hallucinations. A score of 0.5 is assigned when it is difficult to determine whether the response contains hallucinations, a situation that occurs when the resolution of the RS image is too low, and the response does not fully match the ground truth, preventing the expert from making a definitive judgment. A score of 1 indicates that the response is free of hallucinations. For automated scoring, since binary classification is easier than ternary classification and thus likely to yield higher accuracy, we prompt the MLLMs to perform binary classification on whether the answers generated are hallucinated and then convert the results into HF rates.

We calculate the HF rate for each type of QA pair on the basis of the type of hallucination, using the average score for each category:
\begin{equation}
    HF_{T} = \frac{1}{|T|}\sum_{i=1}^{|T|} S_{T_i}
\end{equation}
where $T$ denotes the set of QA pairs corresponding to a specific type of hallucination, $S_{T_i}$ is the score of the $i$-th QA pair within the set, and $HF_{T}$ signifies the HF rate for that type. We use $HF_{IA}$, $HF_{IS}$, $HF_{OE}$, $HF_{OA}$, and $HF_{OR}$ to represent the HF rates in terms of image attributes, image scenes, object existence, object attributes, and object relations, respectively.
The average score of the evaluated model on the RSHalluEval dataset is utilized as the overall HF rate, which measures the model's comprehensive level of hallucination mitigation:
\begin{equation}
    HF_{all} = \frac{1}{|all|}\sum_{i=1}^{|all|} S_{{all}_i}
\end{equation}
where $all$ denotes the set of all QA pairs, $S_{{all}_i}$ is the score of the $i$-th QA pair within the dataset, and $HF_{all}$ signifies the model's overall HF rate.

We also apply the low-hallucination model trained with RSHalluShield to the tasks of RSVQA and RSVG and compare it with state-of-the-art models via mainstream evaluation metrics.

For automated hallucination evaluation methods, we employ evaluation score (ES) to assess their performance, which is calculated based on the relative error between the results of the checker under evaluation and expert assessments. Specifically, given the HF rates of an MLLM on a given type of QA pair, as judged by the expert and by an automated strategy, the corresponding ES is computed as follows:
\begin{equation}
   ES_{T} = \frac{|HF_{T}^{A} - HF_{T}^{E}|}{HF_{T}^{E}}
\end{equation}
where $HF_{T}^{A}$ and $HF_{T}^{E}$ denote the HF rates assessed by the automated evaluation scheme and expert evaluation, respectively. $ES_{T}$ is the ES for this automated evaluation strategy on $T$ for the given MLLM. We use $ES_{IA}$, $ES_{IS}$, $ES_{OE}$, $ES_{OA}$, and $ES_{OR}$ to represent ES in terms of image attributes, image scenes, object existence, object attributes, and object relations, respectively. Similarly, for $HF_{all}$, the corresponding ES is computed as follows:
\begin{equation}
    ES_{all} = \frac{|HF_{all}^{A} - HF_{all}^{E}|}{HF_{all}^{E}}
\end{equation}
where $HF_{all}^{A}$ and $HF_{all}^{E}$ denote the $HF_{all}$ assessed by the automated evaluation scheme and expert evaluation, respectively. $ES_{all}$ is the overall ES of the evaluation scheme for the given MLLM. In this section, we present the mean evaluation score (MES), denoted as $\overline{ES_T}$ and $\overline{ES_{all}}$, which are obtained by averaging the assessment results of all the MLLMs in Table \ref{Etable1} through a certain evaluation method.

For the performance of the checkers on the RSHalluCheck validation dataset, we employ accuracy as the metric, where the accuracies for different question categories and the overall accuracy are denoted as $A_{IA}$, $A_{IS}$, $A_{OE}$, $A_{OA}$, $A_{OR}$, and $A_{all}$, respectively.

\subsection{
Hallucination Benchmarking Results on RSHalluEval
}
\label{sec:VII-C}
The results of evaluating the hallucination performance of MLLMs in the RS domain via the RSHalluEval dataset are shown in Table \ref{Etable1}. The first three models are general-domain MLLMs, whereas the last four are RS-specific MLLMs. \enquote{Expert} denotes expert evaluation, \enquote{Qwen-API-CoT} indicates the evaluation using the API of Qwen-VL-max with the prompt guiding the model to answer in a CoT manner, and \enquote{Qwen2-VL} signifies the evaluation method of the hallucination checker trained on the RSHalluCheck dataset. For the sake of rigor, we rely primarily on the evaluation results of RS experts for the comparison of different MLLMs in Table \ref{Etable1}, and the evaluation results of automated evaluation strategies are utilized to demonstrate their effectiveness in assessment.

As illustrated in Table \ref{Etable1}, among the general-domain models, LLaVA-1.5 manifests the most severe hallucinations, whereas mPLUG-Owl3 presents the least severe hallucinations. Among the RS-specific models, LHRS-Bot results in the most severe hallucinations, and VHM results in the least severe hallucinations. In general, there is a correlation between model performance and the degree of hallucination.

\begin{table}[ht]
    \centering
    \caption{Evaluating the Hallucination of MLLMs on the RSHalluEval Dataset}

    \begin{adjustbox}{width=1\textwidth}  
    
    
    \begin{tabular}{llllllll}  
        \toprule
        \multirow{2}{*}{MLLMs} & \multirow{2}{*}{Assessment Strategies} & \multicolumn{2}{c}{Image Level} & \multicolumn{3}{c}{Object Level} & Overall \\
        \cmidrule(lr){3-4} \cmidrule(lr){5-7} \cmidrule(lr){8-8}
        & & $HF_{IA}\uparrow$ & $HF_{IS}\uparrow$ & $HF_{OE}\uparrow$ & $HF_{OA}\uparrow$ & $HF_{OR}\uparrow$ & $HF_{all}\uparrow$ \\
        \midrule
        \multirow{3}{*}{LLaVA-1.5 \cite{LLaVA}} & Expert & 0.3750 & \underline{0.8462} & 0.5597 & 0.4857 & 0.5357 & 0.5317 \\
        & Qwen-API-CoT & 0.3333 & 0.5750 & 0.6601 & 0.3519 & 0.3679 & 0.5007 \\
        & Qwen2-VL & 0.8542 & 0.9125 & 0.7858 & 0.6034 & 0.4623 & 0.7084 \\
        \midrule
        \multirow{3}{*}{mPLUG-Owl3 \cite{mPLUG-Owl3}} & Expert & 0.1667 & 0.8200 & \textbf{0.8661} & \underline{0.5170} & \underline{0.7500} & \textbf{0.6900} \\
        & Qwen-API-CoT & 0.2778 & 0.7750 & 0.8231 & 0.4846 & 0.5566 & 0.6441 \\
        & Qwen2-VL & 0.6806 & 0.9250 & 0.8428 & 0.7160 & 0.4811 & 0.7593 \\
        \midrule
        \multirow{3}{*}{Qwen2-VL \cite{Qwen-VL}} & Expert & 0.6111 & 0.7941 & 0.7428 & \textbf{0.5745} & 0.6818 & 0.6783 \\
        & Qwen-API-CoT & 0.6181 & 0.7813 & 0.8242 & 0.5864 & 0.5802 & 0.7044 \\
        & Qwen2-VL & 0.9792 & 0.9563 & 0.9022 & 0.8318 & 0.7594 & 0.8744 \\
        \midrule
        \multirow{3}{*}{GeoChat \cite{GeoChat}} & Expert & \underline{0.7500} & 0.5455 & 0.4167 & 0.3750 & 0.4833 & 0.4417 \\
        & Qwen-API-CoT & 0.5625 & 0.3188 & 0.4994 & 0.3025 & 0.2877 & 0.4043 \\
        & Qwen2-VL & 0.8056 & 0.8375 & 0.7404 & 0.6296 & 0.4434 & 0.6861 \\
        \midrule
        \multirow{3}{*}{LHRS-Bot \cite{LHRS-Bot}} & Expert & 0.1667 & 0.4211 & 0.4346 & 0.2898 & 0.4242 & 0.3633 \\
        & Qwen-API-CoT & 0.2431 & 0.3063 & 0.4878 & 0.2731 & 0.3255 & 0.3702 \\
        & Qwen2-VL & 0.7292 & 0.7250 & 0.5995 & 0.4753 & 0.4198 & 0.5601 \\
        \midrule
        \multirow{3}{*}{LHRS-Bot-Nova \cite{LHRS-Bot-Nova}} & Expert & 0.7273 & \textbf{0.8696} & \underline{0.7946} & 0.4853 & 0.5417 & 0.6700 \\
        & Qwen-API-CoT & 0.6736 & 0.6625 & 0.7893 & 0.4475 & 0.4528 & 0.6263 \\
        & \textit{Qwen2-VL} & 0.9167 & 0.8938 & 0.7532 & 0.6975 & 0.4717 & 0.7286 \\
        \midrule
        \multirow{3}{*}{VHM \cite{VHM}} & Expert & \textbf{0.9130} & 0.8125 & 0.7521 & 0.4853 & \textbf{0.8000} & \underline{0.6833} \\
        & Qwen-API-CoT & 0.8958 & 0.7000 & 0.7520 & 0.3611 & 0.4906 & 0.6055 \\
        & Qwen2-VL & 0.9167 & 0.9438 & 0.8137 & 0.6590 & 0.4387 & 0.7425 \\
        \bottomrule[1pt] 
        \multicolumn{8}{l}{\small Bold indicates the best result under a certain metric, while underline indicates the second-best result. We only} \\
        \multicolumn{8}{l}{\small annotate bold and underline the conditions of expert evaluation. }\\
    \end{tabular}
    \end{adjustbox}
    \label{Etable1}
\end{table}

Notably, as a general MLLM, mPLUG-Owl3 is even perform better than that of most of the RS-specific models. This can be attributed to the powerful capabilities of mPLUG-Owl3, which enable it to handle a wide variety of tasks and adapt to new tasks and domains. Furthermore, some of its unique design elements, such as the Hyper Attention Transformer Block, may help enhance the fusion between visual and textual modalities, improving modality consistency and thereby mitigating hallucinations. In contrast, GeoChat and VHM, which are based on LLaVA, may be constrained by the baseline performance. Moreover, constrained by the quantity and distribution of training data, GeoChat may struggle to answer misleading questions effectively.

Furthermore, although mPLUG-Owl3 and Qwen2-VL demonstrate impressive overall performance, they exhibit lower accuracy compared to LHRS-Bot-Nova and VHM on questions related to image attributes. The rationale underlying this observation is that questions pertaining to image attributes in the RS domain exhibit diminished disparities compared with those in the general domain. In contrast, the performance gap in other types of hallucinations between MLLMs in these two domains is not significant. This indicates that image-related hallucinations exhibit specificity within the RS domain, highlighting the necessity of studying image hallucinations in this domain.

\subsection{
Evaluating the Checkers
}
\label{sec:VII-E}
In this section, we conduct a thorough investigation of the effectiveness of online invocation and local deployment strategies, encompassing a variety of prompt strategies for online methods, the categories of models selected for local deployment, and the fine-tuning results. We also compare the accuracy and efficiency of the online and local strategies and provide context-specific recommendations for their application.
\subsubsection{
Online Mode Prompting Ablations
}
Table \ref{Etable2} shows the MES of employing the API under different prompts. We examine two decision-making strategies: one for indicating the presence of hallucinations in the output (1 for presence, 0 for absence) and another for accuracy (1 for without hallucinations and accurate, 0 for with hallucinations and inaccurate), represented in the table as \enquote{Presence} and \enquote{Accuracy}, respectively. We also set up experimental groups without CoT for comparison.

\begin{table}[]
    \centering
    \caption{MESs of Different Prompting Strategies}

    \begin{adjustbox}{width=1\textwidth}  
    \begin{tabular}{llllllllll}
        \toprule
        
        \multirow{2}{*}{CoT} & \multirow{2}{*}{Marking Strategies} & \multirow{2}{*}{Answers} & \multicolumn{2}{c}{Image Level} & \multicolumn{3}{c}{Object Level} & Overall & \multirow{2}{*}{Reference Time (s)} \\
        \cmidrule(lr){4-5} \cmidrule(lr){6-8} \cmidrule(lr){9-9}
        & & & $\overline{ES_{IA}}\downarrow$ & $\overline{ES_{IS}}\downarrow$ & $\overline{ES_{OE}}\downarrow$ & $\overline{ES_{OA}}\downarrow$ & $\overline{ES_{OR}}\downarrow$ & $\overline{ES_{all}}\downarrow$ & \\
        \midrule
        $\checkmark$ & Presence & $\checkmark$ & 0.3257 & 0.2813 & \textbf{0.0617} & 0.1489 & 0.2857 & 0.0996 & - \\
        \ding{55} & Presence & $\checkmark$ & 0.1925 & 0.2541 & 0.0649 & 0.2380 & 0.3480 & 0.1302 & - \\
        $\checkmark$ & Accuracy & $\checkmark$ & 0.2272 & \textbf{0.2081} & 0.0952 & \textbf{0.1348} & 0.2726 & \textbf{0.0637} & - \\
        \ding{55} & Accuracy & $\checkmark$ & \textbf{0.1903} & 0.2545 & 0.0747 & 0.2130 & 0.3299 & 0.1119 & - \\
        \ding{55} & Accuracy & \ding{55} & 0.5945 & 0.2409 & 0.0671 & 0.1572 & \textbf{0.2388} & 0.1156 & 3601.2814 \\
        \bottomrule
    \end{tabular}
    \end{adjustbox}  
    \label{Etable2}
\end{table}

By comparing the $\overline{ES_{all}}$ between the first four rows, it can be concluded that decision marks significantly affect the accuracy. This is probably because marking whether an answer is correct is a more intuitive process. Therefore, we use accuracy marking for strategies with and without using CoT.

In addition, incorporating the guidance to use CoT in the prompt can increase the overall accuracy of the automated evaluation, decreasing $\overline{ES_{all}}$ by 23.50\% and 43.07\% under the marking strategies of \enquote{Presence} and \enquote{Accuracy}, respectively. When the CoT strategy is used, the evaluation results show minimal differences from those of human evaluation, demonstrating that large general-domain models with a substantial number of parameters can accurately evaluate the hallucination status of RS MLLMs through reasonable prompt design.

We also present in the last row of Table \ref{Etable2} the MESs and average reference time required for evaluating MLLMs in Table \ref{Etable1} on the RSHalluEval dataset (denoted as \enquote{Reference Time}) when not using CoT, under the optimal settings of the marking strategy, without using ground truth answers for evaluation, for comparison with Table \ref{Etable4}.

\subsubsection{
Local Checker Results and Efficiency
}
Table \ref{Etable3} presents the performance of the zero-shot and fine-tuned open-source models on the RSHalluCheck validation set. Under zero-shot conditions, MLLMs have relatively low detection accuracy. Among them, LLaVA-1.5 has the lowest overall accuracy, while mPLUG-Owl3 achieves the highest overall accuracy. We then perform fine-tuning of Qwen2-VL and mPLUG-Owl3 on RSHalluCheck. After fine-tuning, the detection accuracy of both models improved significantly, and their overall performance became quite close, with mPLUG-Owl3 outperforming Qwen2-VL by only 1.39\%.

\begin{table}[]
    \centering
    \caption{Precision of Local Hallucination Checkers on the RSHalluCheck Validation Set}

    \begin{adjustbox}{width=1\textwidth}  
    \begin{tabular}{llllllll}
        \toprule
        \multirow{2}{*}{MLLMs} & \multirow{2}{*}{Training Strategies} & \multicolumn{2}{c}{Image Level} & \multicolumn{3}{c}{Object Level} & Overall \\
        \cmidrule(lr){3-4} \cmidrule(lr){5-7} \cmidrule(lr){8-8}
        & & $A_{IA}\uparrow$ & $A_{IS}\uparrow$ & $A_{OE}\uparrow$ & $A_{OA}\uparrow$ & $A_{OR}\uparrow$ & $A_{all}\uparrow$ \\
        \midrule
        LLaVA-1.5 \cite{LLaVA} & zero-shot & 0.7000 & 0.5815 & 0.5433 & 0.5333 & 0.5333 & 0.5521 \\
        \cmidrule(lr){1-8}
        \multirow{2}{*}{Qwen2-VL \cite{Qwen-VL}} & zero-shot & 0.6000 & 0.7120 & 0.6967 & 0.5622 & 0.5676 & 0.6405 \\
        & fine-tuning & 0.6500 & \textbf{0.8913} & 0.9450 & 0.8178 & 0.7432 & 0.8716 \\
        \cmidrule(lr){1-8}
        \multirow{2}{*}{mPLUG-Owl3 \cite{mPLUG-Owl3}} & zero-shot & \textbf{0.8000} & 0.7663 & 0.7883 & 0.6289 & 0.7162 & 0.7268 \\
        & fine-tuning & \textbf{0.8000} & \textbf{0.8913} & \textbf{0.9617} & \textbf{0.8222} & \textbf{0.7568} & \textbf{0.8837} \\
        \bottomrule
    \end{tabular}
    \end{adjustbox}  
    \label{Etable3}
\end{table}

To further demonstrate the hallucination detection performance of the trained checkers in practical applications, we provide the MESs and average reference time required for evaluating MLLMs (on an RTX 4090 GPU with 24GB memory) in Table \ref{Etable1} of the fine-tuned Qwen2-VL and mPLUG-Owl3 on the RSHalluEval dataset, as shown in Table \ref{Etable4}.
\begin{table*}[!h]
    \centering
    \caption{MESs of Fine-tuned Local Hallucination Checkers}

    
    \begin{adjustbox}{width=1\textwidth}  
    \begin{tabular}{lllllllll}
        \toprule
        \multirow{2}{*}{MLLMs} & \multicolumn{2}{c}{Image Level} & \multicolumn{3}{c}{Object Level} & Overall & \multirow{2}{*}{Reference Time (s)} \\
        \cmidrule(lr){2-3} \cmidrule(lr){4-6} \cmidrule(lr){7-7}
        & ${\overline{ES_{IA}}\downarrow}$ & ${\overline{ES_{IS}}\downarrow}$ & ${\overline{ES_{OE}}\downarrow}$ & ${\overline{ES_{OA}}\downarrow}$ & ${\overline{ES_{OR}}\downarrow}$ & ${\overline{ES_{all}}\downarrow}$ & \\
        \midrule
        Qwen2-VL \cite{Qwen-VL} & 1.2396 & \textbf{0.2653} & \textbf{0.2766} & \textbf{0.4557} & \textbf{0.1833} & \textbf{0.2844} & 266.4586 \\
        mPLUG-Owl3 \cite{mPLUG-Owl3} & \textbf{0.8277} & 0.3084 & 0.3303 & 0.6019 & 0.2053 & 0.3598 & 351.3557 \\
        \bottomrule
    \end{tabular}
    \end{adjustbox}  
    \label{Etable4}
\end{table*}

It can be seen from Table \ref{Etable4} that there is little difference between local deployment and online invocation strategies in terms of image scenes, and even local deployment achieves higher accuracy in object relation questions. However, the local deployment strategies significantly lag behind the online invocation strategies in terms of image attributes, object existence, and object attributes. Nevertheless, the inference speeds of the local deployment strategies are significantly faster than those of the online invocation strategy. Consequently, it is recommended to employ local deployment strategies when efficiency is a priority and to utilize the online invocation strategy when emphasizing evaluation accuracy, particularly for the accuracy of each type of hallucination.

Most of the evaluation results of Qwen2-VL are better than those of mPLUG-Owl3 in Table \ref{Etable4}, leading us to employ Qwen2-VL as the local hallucination checker.

\subsection{
Evaluating Training-Free Mitigation Methods
}
\label{sec:VII-F}

In this section, we validate the proposed decoding-based strategy, assess the rationality of reference layer selection, and confirm the effectiveness of the prompt-based decoding strategy.

\subsubsection{
Decoding-based Mitigation
}
In this section, we select one general-domain MLLM, LLaVA-1.5, as well as two RS MLLMs, GeoChat and VHM, to verify the effectiveness of the decoding-based hallucination mitigation scheme. The performance of these MLLMs is presented in Table \ref{Etable12}. In the table, \enquote{Baseline} denotes the experimental groups that do not employ the decoding method, while \enquote{Decoding} refers to the groups that adopt the scheme, and \enquote{Average} denotes the groups in which layers are randomly selected from $M_{origin}$ and weighted.

As illustrated in Table \ref{Etable12}, the proposed method effectively mitigates hallucination across multiple models, achieving the most substantial improvement on GeoChat, where the overall hallucination reduction rate increased by 6.35\% and 6.48\% under greedy and sampling modes, respectively. Notably, the hallucination mitigation effect is particularly prominent for object existence questions, as the proposed decoding strategy consistently enhances the hallucination reduction rate for such questions across all tested models and decoding configurations. This phenomenon likely stems from the method's superior probability calibration capability in binary-choice scenarios (e.g., yes/no determinations), given that object existence answers are inherently binary in nature.

However, in practice, methods that optimize outputs by averaging or weighted summation of results from multiple intermediate layers of MLLMs are widespread. To investigate whether the accuracy improvement stems from the intermediate layer selection strategy or is merely an incidental effect of simple weighted summation across multiple layers, we design controlled experiments. Specifically, we simulate the scenario of simple weighted summation of intermediate layers by randomly selecting $K_m$ layers from the reference layers, averaging their logits, and applying weighted correction with the same backtracking rate $r$ under the sampling search configuration. The results corresponding to this simulation setting are labeled as \enquote{Average} in Table \ref{Etable12}.

As shown in the table, random selection from the reference layers yields limited improvement of the HF rates, even exhibiting suboptimal performance in comparison to the baseline. This finding serves to substantiate the hypothesis that the effect in hallucination reduction achieved by the proposed method stems from the designed intermediate layer selection strategy.

\begin{table*}[]
    \centering
    \caption{Hallucination State of MLLMs Before and After Applying Decoding-based Hallucination Mitigation Scheme}
    \begin{adjustbox}{width=1\textwidth}  
    \begin{tabular}{lllllllll}
        \toprule
        \multirow{2}{*}{MLLMs} & \multirow{2}{*}{Decoding Strategies} & \multirow{2}{*}{Settings} & \multicolumn{2}{c}{Image Level} & \multicolumn{3}{c}{Object Level} & Overall \\
        \cmidrule(lr){4-5} \cmidrule(lr){6-8} \cmidrule(lr){9-9}
        & & & ${HF_{IA}\uparrow}$ & ${HF_{IS}\uparrow}$ & ${HF_{OE}\uparrow}$ & ${HF_{OA}\uparrow}$ & ${HF_{OR}\uparrow}$ & {${HF_{all}\uparrow}$} \\
        \midrule
        \multirow{5}{*}{LLaVA-1.5 \cite{LLaVA}} & \multirow{2}{*}{Greedy} & Baseline & \textbf{0.3125} & \textbf{0.6063} & 0.6449 & 0.3565 & 0.3679 & 0.4968 \\
        & & Decoding (ours) & 0.2014 & 0.5375 & \textbf{0.6822} & \textbf{0.3580} & \textbf{0.4057} & \textbf{0.5037} \\
        \cmidrule(lr){2-9}
        & \multirow{3}{*}{Sampling} & Baseline & \textbf{0.3125} & 0.5313 & 0.6589 & 0.3472 & 0.3632 & 0.4933 \\
        & & Average & 0.2222 & 0.5375 & 0.6449 & 0.3503 & 0.3443 & 0.4805 \\
        & & Decoding (ours) & 0.2153 & \textbf{0.5563} & \textbf{0.6775} & \textbf{0.3627} & \textbf{0.3868} & \textbf{0.5037} \\
        \midrule
        \multirow{5}{*}{GeoChat \cite{GeoChat}} & \multirow{2}{*}{Greedy} & Baseline & 0.5625 & 0.2938 & 0.5029 & \textbf{0.3056} & \textbf{0.2877} & 0.4048 \\
        & & Decoding (ours) & \textbf{0.6111} & \textbf{0.3313} & \textbf{0.5576} & 0.3025 & 0.2594 & \textbf{0.4305} \\
        \cmidrule(lr){2-9}
        & \multirow{3}{*}{Sampling} & Baseline & 0.5625 & 0.2750 & 0.5099 & 0.3040 & \textbf{0.2736} & 0.4043 \\
        & & Average & 0.5347 & 0.2500 & 0.5146 & 0.3272 & 0.2642 & 0.4088 \\
        & & Decoding (ours) & \textbf{0.5972} & \textbf{0.2938} & \textbf{0.5611} & \textbf{0.3086} & 0.2642 & \textbf{0.4305} \\
        \midrule
        \multirow{5}{*}{VHM \cite{VHM}} & \multirow{2}{*}{Greedy} & Baseline & 0.9028 & 0.6938 & 0.7602 & 0.3549 & 0.5142 & 0.6095 \\
        & & Decoding (ours) & \textbf{0.9097} & \textbf{0.7125} & \textbf{0.7846} & \textbf{0.3580} & \textbf{0.5377} & \textbf{0.6253} \\
        \cmidrule(lr){2-9}
        & \multirow{3}{*}{Sampling} & Baseline & \textbf{0.9167} & 0.6875 & \textbf{0.7614} & 0.3673 & 0.5236 & 0.6154 \\
        & & Average & 0.8403 & 0.6188 & 0.7660 & 0.3441 & \textbf{0.5094} & 0.5976 \\
        & & Decoding (ours) & 0.8958 & \textbf{0.7438} & \textbf{0.7614} & \textbf{0.3796} & \textbf{0.5094} & \textbf{0.6209} \\
        \bottomrule
    \end{tabular}
    \end{adjustbox}  
    \label{Etable12}
\end{table*}

\begin{figure}[h!]
\centering
\includegraphics[width=0.8\textwidth]{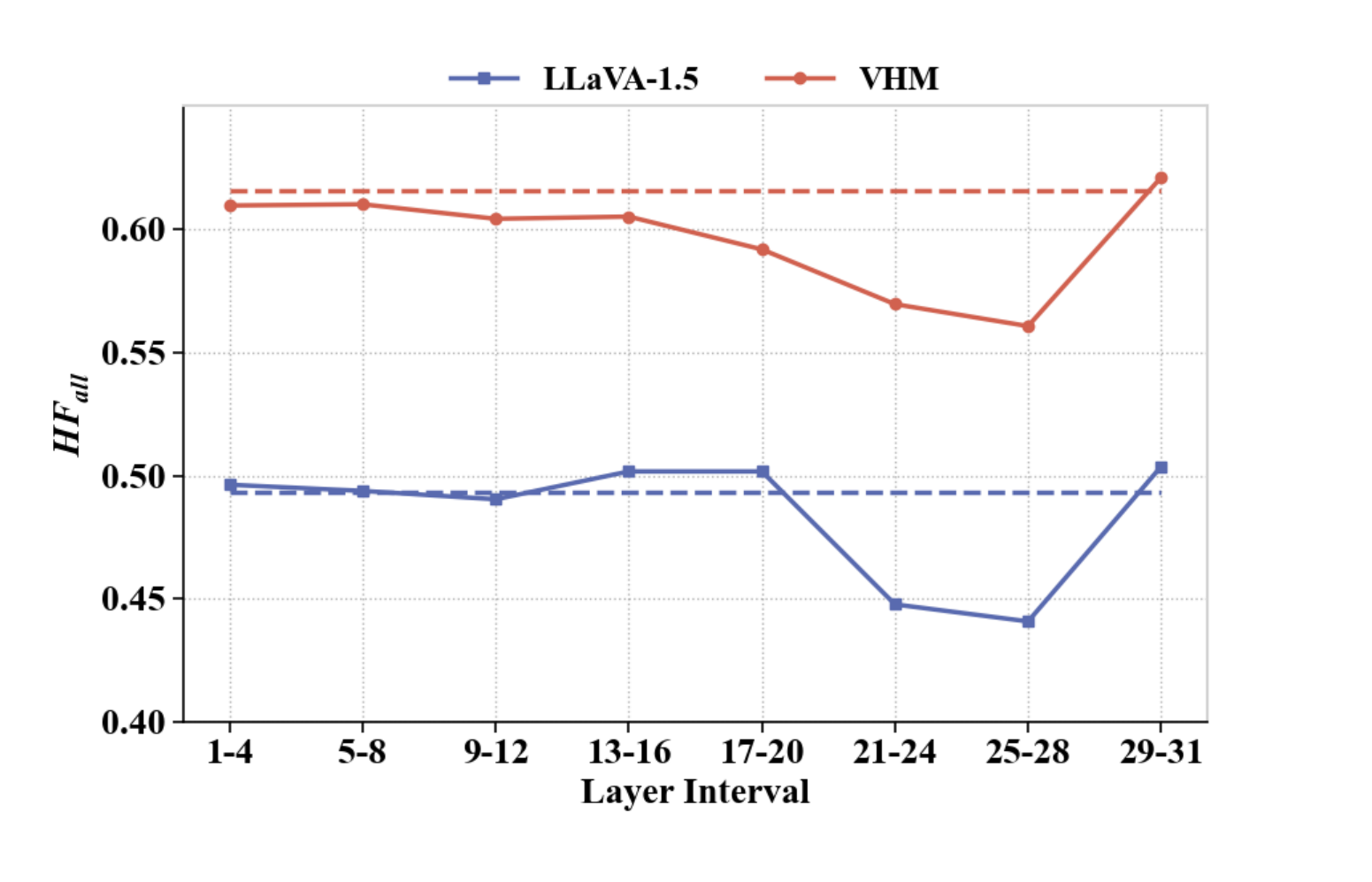}
\caption{The effect of different ranges of $M_{origin}$ on the degree of hallucinations.}
\label{decoder_diff_layer_interval}
\end{figure}

Furthermore, we conduct experiments on the selection of the range of $M_{origin}$, with the results presented in Fig. \ref{decoder_diff_layer_interval}. We conduct the experiments on VHM and LLaVA-1.5, adopting sampling decoding. In the figure, VHM and LLaVA-1.5 exhibit similar trends in their HF rates with varying ranges of reference layers. When layers 1–20 are selected as reference layers, the correction effect yields limited improvement in hallucination mitigation. Choosing layers 21–28 as reference layers leads to a significant decrease in HF rates. We attribute this to the fact that these layers tend to prioritize linguistic priors over visual semantic information. However, selecting layers 29–31 as reference layers mitigates hallucinations for both models. Thus, we set the range of $M_{origin}$ to 29–31.

\subsubsection{
Prompt-based Mitigation
}
We utilize LLaVA-1.5, Qwen2-VL, and VHM to verify the effectiveness of the prompt-based hallucination mitigation schemes. We apply the counterfactual perception strategy and the overall perception strategy to these three MLLMs and study the effect of combining the two strategies, as shown in Table \ref{Etable11}. The above three strategies are represented by \enquote{Counterfactual}, \enquote{Overall} and \enquote{Combined}, respectively, and \enquote{None} indicates that no strategy is used.

As seen from the table, the counterfactual perception strategy effectively improves the overall performance of each MLLM by 13.82\%, 5.89\% and 8.49\% on LLaVA-1.5, Qwen2-VL, and VHM, respectively. This suggests that MLLMs generally struggle with counterfactual input and that certain mitigation effects can be achieved simply by optimizing the prompt.

However, the impact of the overall perception strategy on model performance varies. After utilization, the general performance of LLaVA-1.5 declined, whereas the performance of Qwen2-VL and VHM marginally increased. This may be due to the relatively weak visual extraction and anti-interference abilities of LLaVA-1.5, which make it vulnerable to interference from irrelevant information in the additional visual description text.

Integrating the two strategies significantly improves the comprehensive performance of LLaVA-1.5, with an observed increase of 21.63\% in its $HF_{all}$ score, thereby achieving optimal results. This can be attributed to the interaction of the two strategies, which enables LLaVA-1.5 to enhance the visual perception ability while ignoring the counterfactual information and the interference information of visual description text. However, combining the two strategies does not significantly improve the overall performance of Qwen2-VL. Furthermore, for VHM, using the counterfactual perception strategy alone can achieve slightly greater comprehensive performance than combining the two strategies. Therefore, we recommend using both strategies simultaneously for MLLMs with lower basic performance and the counterfactual perception strategy for those with higher performance.

\begin{table*}[]
    \centering
    \caption{Hallucination State of MLLMs Before and After Applying Prompting Strategies}
    \begin{adjustbox}{width=1\textwidth}  
    \begin{tabular}{llllllll}
        \toprule
        \multirow{2}{*}{MLLMs} & \multirow{2}{*}{Prompting Strategies} & \multicolumn{2}{c}{Image Level} & \multicolumn{3}{c}{Object Level} & Overall \\
        \cmidrule(lr){3-4} \cmidrule(lr){5-7} \cmidrule(lr){8-8}
        & & ${HF_{IA}}\uparrow$ & ${HF_{IS}}\uparrow$ & ${HF_{OE}}\uparrow$ & ${HF_{OA}}\uparrow$ & ${HF_{OR}}\uparrow$ & {${HF_{all}\uparrow}$} \\
        \midrule
        \multirow{4}{*}{LLaVA-1.5 \cite{LLaVA}} & None & 0.3333 & 0.5750 & 0.6601 & 0.3519 & 0.3679 & 0.5007 \\
        & Counterfactual & \textbf{0.5694} & 0.5125 & 0.7474 & \textbf{0.4074} & 0.3915 & 0.5699 \\
        & Overall & 0.3056 & \textbf{0.6000} & 0.6123 & 0.3657 & 0.3019 & 0.4780 \\
        & Combined & 0.5278 & 0.5625 & \textbf{0.8289} & 0.4059 & \textbf{0.4292} & \textbf{0.6090} \\
        \cmidrule(lr){1-8}
        \multirow{4}{*}{Qwen2-VL \cite{Qwen-VL}} & None & 0.6181 & 0.7813 & 0.8242 & 0.5864 & 0.5802 & 0.7044 \\
        & Counterfactual & 0.6667 & \textbf{0.8625} & \textbf{0.8696} & \textbf{0.6127} & 0.6179 & 0.7459 \\
        & Overall & \textbf{0.7986} & 0.7875 & 0.8382 & 0.5370 & 0.5708 & 0.7069 \\
        & Combined & 0.7847 & 0.8188 & 0.8661 & 0.5880 & \textbf{0.6651} & \textbf{0.7464} \\
        \cmidrule(lr){1-8}
        \multirow{4}{*}{VHM \cite{VHM}} & None & 0.8958 & 0.7000 & 0.7520 & 0.3611 & 0.4906 & 0.6055 \\
        & Counterfactual & \textbf{0.9375} & \textbf{0.7813} & \textbf{0.7730} & \textbf{0.4676} & 0.4811 & \textbf{0.6569} \\
        & Overall & \textbf{0.9375} & 0.6688 & 0.7218 & 0.4383 & \textbf{0.5142} & 0.6204 \\
        & Combined & 0.9306 & 0.7375 & 0.7707 & 0.4599 & 0.5000 & 0.6515 \\
        \bottomrule
        
    \end{tabular}
    \end{adjustbox}  
    \label{Etable11}
\end{table*}

\subsection{
RSHalluShield Design Analysis
}
\label{sec:VII-G}
In this section, we first demonstrate the effectiveness of the RSHalluShield dataset in mitigating hallucinations. The ablation study is subsequently conducted to validate the scientific rationale behind its design.

\subsubsection{
Effectiveness of the RSHalluShield
}
Table \ref{Etable5} illustrates the performance differences of Qwen2-VL before and after fine-tuning (denoted as Qwen2-VL-HF in this work) with RSHalluShield. Specifically, the degree of hallucination in Qwen2-VL decreases, accompanied by 46.06\%, 6.36\%, 6.58\%, and 23.58\% increases in the HF rates in terms of image attributes, object existence, object attributes, and object relations, respectively. However, the degree of hallucinations decreases significantly in the categories of image scenes. This might be because RSHalluShield is automatically annotated, where the answers to such questions usually consist of a single sentence describing the image scene. In RSHalluEval, however, expert annotations for these types of questions often consist of a single word representing the scene's category. Consequently, the responses of the fine-tuned model in this type of question differ from the given answers.

\begin{table*}[ht]
    \centering
    \caption{Hallucination State of Qwen2-VL Before and After Fine-tuning}
    \begin{adjustbox}{width=0.7\textwidth}  
    \begin{tabular}{lllllll}
        \toprule
        \multirow{2}{*}{MLLMs} & \multicolumn{2}{c}{Image Level} & \multicolumn{3}{c}{Object Level} & Overall \\
        \cmidrule(lr){2-3} \cmidrule(lr){4-6} \cmidrule{7-7}
        & ${HF_{IA}}\uparrow$ & ${HF_{IS}}\uparrow$ & ${HF_{OE}}\uparrow$ & ${HF_{OA}}\uparrow$ & ${HF_{OR}}\uparrow$ & {${HF_{all}\uparrow}$} \\

        \midrule
        Qwen2-VL & 0.6181 & \textbf{0.7813} & 0.8242 & 0.5864 & 0.5802 & 0.7044 \\
        Qwen2-VL-HF & \textbf{0.9028} & 0.7750 & \textbf{0.8766} & \textbf{0.6250} & \textbf{0.7170} & \textbf{0.7731} \\
        \bottomrule
    \end{tabular}
    \end{adjustbox}  
    \label{Etable5}
\end{table*}

\subsubsection{
Ratio Study in the RSHalluShield
}
LRV-Instruction employs an equal number of normal and misleading examples during training; however, this ratio is not directly applied in our work. On the one hand, there are domain differences between the RS and general domains. On the other hand, the proportion of misleading questions to nonmisleading questions in real-world scenarios may not be balanced; it is likely that nonmisleading questions predominate. In the context of the RSHalluEval dataset, there are still more nonmisleading questions than misleading ones. Under such circumstances, including a balanced number of normal and misleading examples in the training set may not lead to the optimal hallucination mitigation effect.

Therefore, we keep the total number of QA pairs unchanged while altering the ratio of normal to misleading examples to investigate the optimal ratio corresponding to the best hallucination mitigation effect. For each trial, we randomly sample a fixed number of QA pairs from pools of different question types to form the training set.

The evaluation results of the models trained on datasets with different ratios on the RSHalluEval dataset are shown in Table \ref{Etable6}. In the table, \enquote{Ratios} indicates the ratio of normal QA pairs to misleading QA pairs. As demonstrated in the table, even though the ratios of normal to misleading examples in the RSHalluEval dataset and the validation set of LRV-Instruction differ, the group with a balanced ratio still achieves the highest HF rate. Furthermore, $HF_{all}$ demonstrate minimal variation, with the smallest difference being 0.0005 and the largest being 0.0128. These findings suggest that the hallucination mitigation effect is relatively insensitive to alterations in this ratio.

\begin{table}[ht]
    \centering
    \caption{The Impact of the Normal and Misleading Ratios on the Degree of Hallucinations}
    \begin{adjustbox}{width=0.7\textwidth}  
    \begin{tabular}{llllllll}
        \toprule
        \multirow{2}{*}{{Ratios}} & \multicolumn{2}{c}{{Image Level}} & \multicolumn{3}{c}{{Object Level}} & {Overall} \\
        \cmidrule(lr){2-3} \cmidrule(lr){4-6} \cmidrule{7-7}
        & ${HF_{IA}}\uparrow$ & ${HF_{IS}}\uparrow$ & ${HF_{OE}}\uparrow$ & ${HF_{OA}}\uparrow$ & ${HF_{OR}}\uparrow$ & {${HF_{all}\uparrow}$} \\
        \midrule
        6 : 9 & 0.8958 & 0.7125 & \textbf{0.8778} & 0.6049 & 0.7406 & 0.7642 \\
        7 : 8 & \textbf{0.9444} & 0.6938 & 0.8545 & 0.6204 & 0.7311 & 0.7603 \\
        1 : 1 & 0.9028 & \textbf{0.7750} & 0.8766 & \textbf{0.6250} & 0.7170 & \textbf{0.7731} \\
        8 : 7 & 0.9097 & 0.7313 & 0.8696 & 0.6003 & \textbf{0.7594} & 0.7637 \\
        \bottomrule
    \end{tabular}
    \end{adjustbox}  
    \label{Etable6}
\end{table}

\subsection{
Downstream Transfer: RSVQA and RSVG
}
\label{sec:VII-H}
To evaluate downstream transferability, we apply Qwen2-VL fine-tuned on RSHalluShield to two representative RS tasks: RSVQA and RSVG.

\subsubsection{
Applied to RSVQA
}
We evaluate the performance of the Qwen2-VL fine-tuned on the RSHalluShield dataset on the RSVQA task via the RSVQA-LR and RSVQA-HR datasets \cite{RSVQA}. RSVQA-LR consists of 772 images with 77,232 QA pairs, which are divided into a training set (77.8\%), a validation set (11.1\%), and a test set (11.1\%). On the other hand, RSVQA-HR contains 10,659 images. The dataset contains 1,066,316 QA pairs, which are divided into a training set (61.5\% of the tiles), a validation set (11.2\%), and a test set (test set 1 accounts for 20.5\% and test set 2 for 6.8\%). We fine-tune the original and low-hallucination Qwen2-VL on RSVQA-LR and perform a zero-shot evaluation on RSVQA-HR, with the results shown in Tables \ref{Etable8} and \ref{Etable9}, respectively.

Table \ref{Etable8} shows that, after fine-tuning on the RSHalluShield dataset, Qwen2-VL-HF achieves state-of-the-art performance on some metrics of RSVQA-LR, outperforming the second place by 2.91 and 0.22 on LR-rural and Avg., respectively, and ranking second on LR-presence and LR-compare, thereby demonstrating the effectiveness of the fine-tuned low-hallucination model. Moreover, Table \ref{Etable9} shows that Qwen2-VL-HF achieves second place for HR-presence and Avg. on RSVQA-HR, which again shows that pretraining on the RSHalluShield dataset can effectively adapt to other RSVQA tasks.

Furthermore, the performance difference between Qwen2-VL and Qwen2-VL-HF here is greater than that in Table \ref{Etable8}. In addition to the differences in dataset features, the fine-tuning on the RSVQA-LR task enables Qwen2-VL to adapt to the RS field, which is one reason for reducing the differences between the two models in Table \ref{Etable8}. However, this also indicates that fine-tuning using the RSHalluShield dataset does not merely enable domain adaptation for MLLMs in the general domain.

\begin{table*}[!ht]
    \centering
    \caption{Quantitative Comparison of the Performance of MLLMs on the RSVQA-LR Dataset}
    \begin{adjustbox}{width=0.7\textwidth}  
    \begin{tabular}{lllll}
        \toprule
        {MLLMs} & {LR-rural} & {LR-presence} & {LR-compare} & {Avg.} \\
        \midrule
        Gemini-Vision \cite{Gemini-Vision} & 63.00 & 60.95 & 70.32 & 64.76 \\
        RSGPT \cite{RSGPT} & \textbf{94.00} & 91.17 & 92.29 & \underline{92.49} \\
        GeoChat \cite{GeoChat} & \underline{91.09} & 90.33 & \textbf{94.00} & 91.81 \\
        SkyEyeGPT \cite{SkyEyeGPT} & 88.93 & 88.63 & 75.00 & 84.19 \\
        LHRS-Bot \cite{LHRS-Bot} & 89.07 & 88.51 & 90.00 & 89.19 \\
        VHM \cite{VHM} & 88.00 & 90.11 & 89.89 & 89.33 \\
        Qwen2-VL \cite{Qwen-VL} & \textbf{94.00} & \textbf{91.91} & 92.20 & 92.11 \\
        Qwen2-VL-HF & \textbf{94.00} & \underline{91.81} & \underline{92.33} & \textbf{92.71} \\
        \bottomrule
    \end{tabular}
    \end{adjustbox}  
    \label{Etable8}
\end{table*}

\begin{table}[!ht]
    \centering
    \caption{Quantitative Comparison of the Performance of MLLMs on the RSVQA-HR Dataset}
    \begin{adjustbox}{width=0.6\textwidth}  
    \begin{tabular}{lllll}
        \toprule
        {MLLMs} & {HR-presence} & {HR-compare} & {Avg.} \\
        \midrule
        Gemini-Vision \cite{Gemini-Vision} & 63.60 & 64.60 & 64.10 \\
        LLaVA-1.5 \cite{LLaVA} & \textbf{69.83} & 67.29 & 68.56 \\
        MiniGPTv2 \cite{MiniGPTv2} & 40.79 & 50.91 & 45.85 \\
        Qwen-VL-Chat \cite{Qwen-VL} & 66.44 & 60.41 & 63.43 \\
        GeoChat \cite{GeoChat} & 58.45 & \underline{83.19} & 70.82 \\
        EarthGPT \cite{EarthGPT} & 62.77 & 79.53 & 71.15 \\
        VHM \cite{VHM} & 64.00 & \textbf{83.50} & \textbf{73.75} \\
        Qwen2-VL \cite{Qwen-VL} & 67.08 & 70.29 & 68.88 \\
        Qwen2-VL-HF & \underline{67.63} & 75.52 & \underline{71.58} \\
        \bottomrule
    \end{tabular}
    \end{adjustbox}  
    \label{Etable9}
\end{table}

\subsubsection{
Applied to RSVG
}
We evaluate the performance of our method on the RSVG task via the DIOR-RSVG \cite{DIOR-RSVG} dataset. The DIOR-RSVG dataset consists of 17,402 RS images and 38,320 linguistic expressions. For the RSVG task, precision is used as the evaluation criterion. A prediction is considered accurate if the intersection over union (IoU) between the predicted bounding box and the ground truth bounding box for a given annotation is greater than 0.5.

Table \ref{Etable10} shows the performance of the MLLMs after being fine-tuned on the DIOR-RSVG dataset. Among them, Qwen2-VL, Qwen-VL-Chat and CogVLM are general domain MLLMs, MGVLF is an RS RSVG model, and VHM is an RS MLLM. Although the RSHalluShield dataset does not include the RSVG task, Qwen2-VL-HF still achieves an accuracy improvement of 0.83 compared with that of Qwen2-VL and outperforms the aforementioned models. This once again demonstrates the role of RSHalluShield in enhancing the overall performance of MLLMs in RS applications.

To further explore the reasons why fine-tuning with RSHalluShield leads to better performance on RSVG tasks, we investigate Qwen2-VL trained using only the image-level QA pairs in RSHalluShield. We present the fine-tuning results on DIOR-RSVG (denoted as Qwen2-VL-w/o-i) in Table \ref{Etable10}. The Qwen2-VL-w/o-i model is the lowest among the three Qwen2-VL models. Consequently, the enhanced performance of Qwen2-VL-HF in RSVG tasks is predominantly attributable to the incorporation of object-level questions within the RSHalluShield framework. These questions inquire about information such as the color, size, and spatial relations of objects, prompting the model to enhance its ability to extract fine-grained features.


\begin{table}[!ht]
    \centering
    \caption{Quantitative Comparison of the Performance of MLLMs on the DIOR-RSVG Dataset}
    \begin{adjustbox}{width=0.47\textwidth}  
    \begin{tabular}{llllll}
        \toprule
        {MLLMs} & {Precision (IoU$>$0.5)} \\
        \midrule
        Qwen-VL-Chat \cite{Qwen-VL} & 31.86 \\
        CogVLM \cite{CogVLM} & 44.58 \\
        MGVLF \cite{MGVLF} & 76.78 \\
        VHM \cite{VHM} & 56.17 \\
        \midrule
        Qwen2-VL \cite{Qwen-VL} & 77.21 \\
        Qwen2-VL-w/o-i & 77.04 \\
        Qwen2-VL-HF & \textbf{78.04} \\
        \bottomrule
    \end{tabular}
    \end{adjustbox}  
    \label{Etable10}
\end{table}

\subsection{Qualitative Results}
\label{sec:VII-I}

To intuitively demonstrate the effectiveness of the proposed method, we present and analyze visual examples from both hallucination evaluation methods and mitigation strategies.

\subsubsection{
Qualitative Analysis of Automated Evaluation
}
\begin{figure}[t!]
\vspace{-3cm}  
\centering
\includegraphics[width=0.88\textwidth]{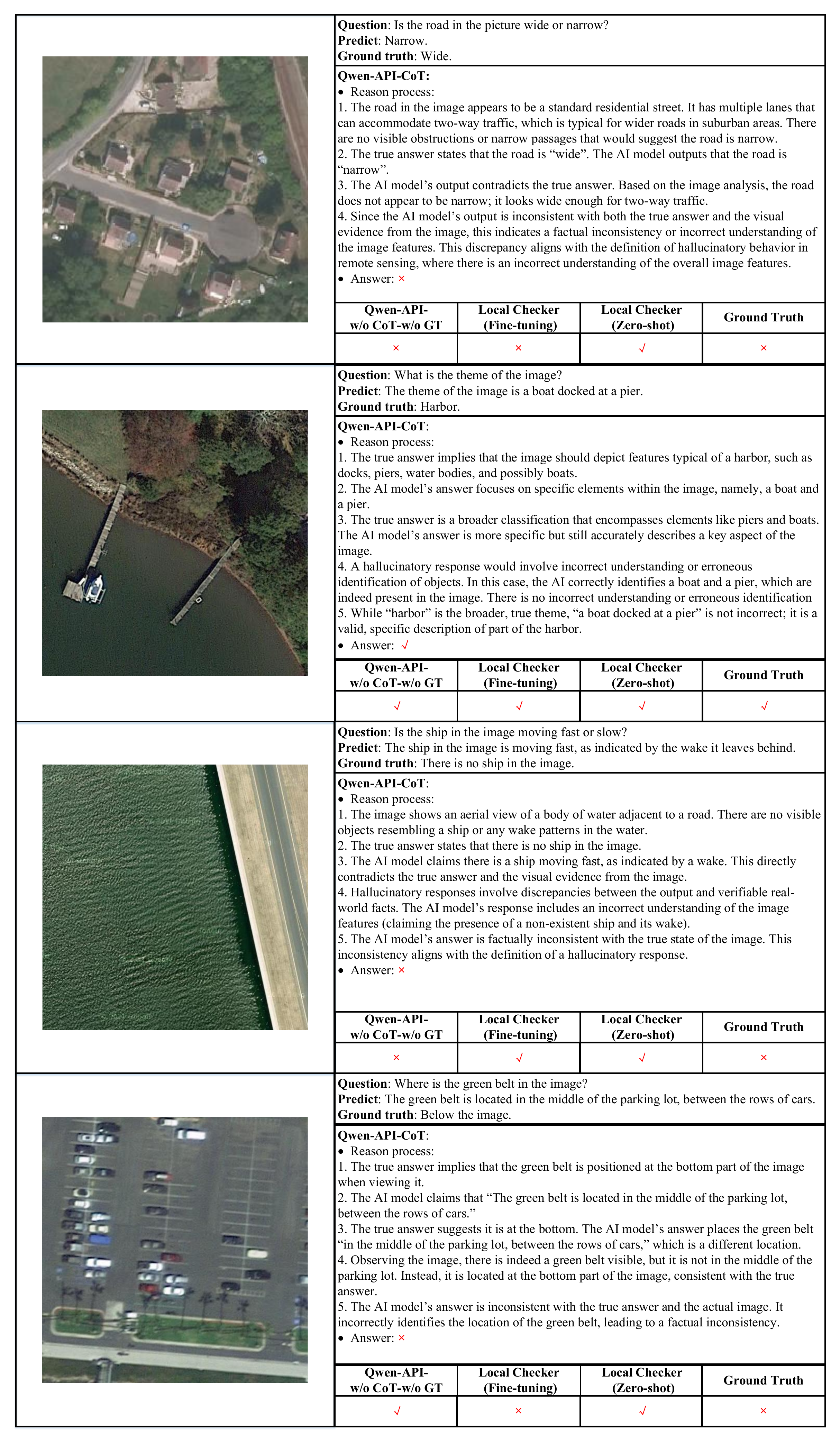}
\caption{Some evaluation results on the RSHalluEval dataset.}
\label{visualize1}
\end{figure}

To visually demonstrate the effectiveness of the proposed automated evaluation strategies, we present several examples of evaluations on the RSHalluEval dataset, as shown in Fig. \ref{visualize1}, with \enquote{Question}, \enquote{Ground truth} denoting the question and answer of the QA pair, and \enquote{Predict} representing the generated text of the MLLM. \enquote{Qwen-API-CoT} and \enquote{Qwen-API-w/o-CoT-w/o-GT} represent the evaluations using the Qwen-VL-Max API with and without the CoT method and ground truth, respectively, whereas \enquote{Local Checker (fine-tuning)} and \enquote{Local Checker (zero-shot)} denote local hallucination checkers with and without fine-tuning, respectively. In the lower right corner of each example, the correct judgment is provided, labelled \enquote{Ground Truth}. In these examples, the response of using CoT to activate Qwen-VL-Max comprises the checker's reasoning processes and judgments, whereas the other includes only decision results. For the decision results, we mark situations without hallucinations with a \enquote{$\checkmark$} and those with hallucinations with a \enquote{\ding{55}}.

It can be seen from Fig. \ref{visualize1} that the local checker without fine-tuning has inaccuracies in determining some simple questions. In the first example, while other evaluation strategies can directly infer the contradiction between the image and the MLLM response, the local checker without fine-tuning fails to make the correct judgment. In fact, it has been observed that there is a tendency to assume that MLLMs do not experience hallucinations. This may be due to the limited parameter size and lack of training data, leading to a biased understanding of hallucinations.

In contrast, the fine-tuned local checker grasps the concept of hallucinations and the methods of judgment. However, it still lags behind API-based strategies in some scenarios. In the third example, the MLLM's answer \enquote{The ship in the image is moving fast, as indicated by the wake it leaves behind.} does not match the true state of the image. The fine-tuned local checker fails to correctly detect the hallucination in the response, possibly because its reasoning ability is limited, and it is also misled by the question to produce hallucinations, thus leading to a misjudgment of the existence of the ship. In comparison, it can be observed from the inference process that Qwen-API-CoT identifies inaccurate information in the generated text on the basis of the input image and the ground truth.

For API-based strategies, the employment of CoT and the reference of ground truth allow the checker to accurately assess the validity of generated text, thereby increasing the applicability to open-ended question answering. In the fourth example, the generated text references the facts present in the image but is hallucinatory. In the absence of CoT application and when the ground truth is not provided, Qwen-VL-Max considers the result to be hallucination-free. However, through a step-by-step analysis, Qwen-API-CoT identifies discrepancies by comparing the ground truth with the predicted answers, observing the images, and concluding that there is a hallucination in the predicted answer.

Although Qwen-VL-Max and the fine-tuned local checker may not achieve perfect accuracy universally, they still exhibit notable capabilities in visual feature extraction and reasoning. In the second example, the predicted answer is more specific, whereas the ground truth is a more concise summary of the entire scene. In this scenario, Qwen-API-CoT can make correct judgments by comparing the predicted answer with the ground truth and the image. Moreover, hallucination checkers that do not reference the ground truth can also correctly judge by extracting the visual features of \enquote{boat} and \enquote{pier}.

\subsubsection{
Qualitative Analysis of Mitigation Effects
}
\begin{figure*}[]
\centering
\includegraphics[width=1\textwidth]{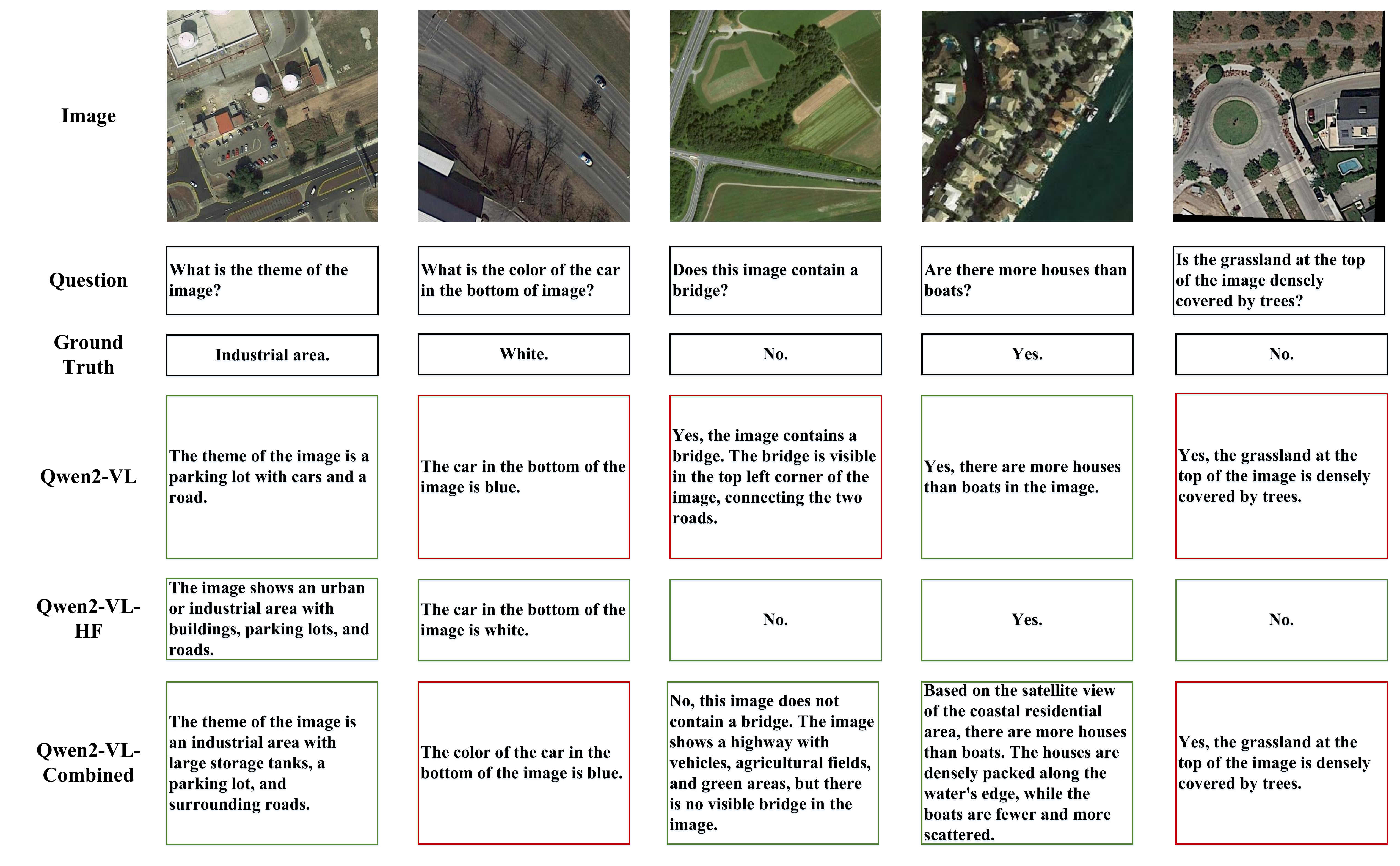}
\caption{Some examples of the performance of Qwen2-VL on RSHalluEval before and after fine-tuning or using the combination of the prompt-based hallucination mitigation schemes. Red boxes indicate hallucinatory answers, whereas green boxes indicate non-hallucinatory answers.}
\label{visualize2}
\end{figure*}

Fig. \ref{visualize2} shows the performance of Qwen2-VL before and after fine-tuning on RSHalluShield or using the combination of the two prompt-based hallucination mitigation schemes, as evaluated on RSHalluEval.

Without any measures to alleviate hallucinations, Qwen2-VL is capable of basic perception and reasoning, such as determining the elements in the image (the first example on the left) and the relation (the fourth example on the left) between houses and cars. After applying prompt-based hallucination mitigation schemes, Qwen2-VL judges the existence of objects more accurately (the third examples from the left). These schemes also enable Qwen2-VL-Combined to draw on RS-related knowledge for more professional responses. For example, in the first example on the left, it recognizes more objects and identifies the land use type in RS images as \enquote{industrial area}, which is closer to the ground truth.

Compared with zero-shot models, Qwen2-VL-HF can better alleviate hallucinations. In the second example from the left, the models without fine-tuning hallucinate about the car's color, yet Qwen2-VL-HF makes a correct judgment. Moreover, Qwen2-VL-HF better understands the semantic information of RS images and accurately assesses the reasonableness of \enquote{densely covered} in the rightmost example.

\begin{figure}[]
\centering
\includegraphics[width=0.75\textwidth]{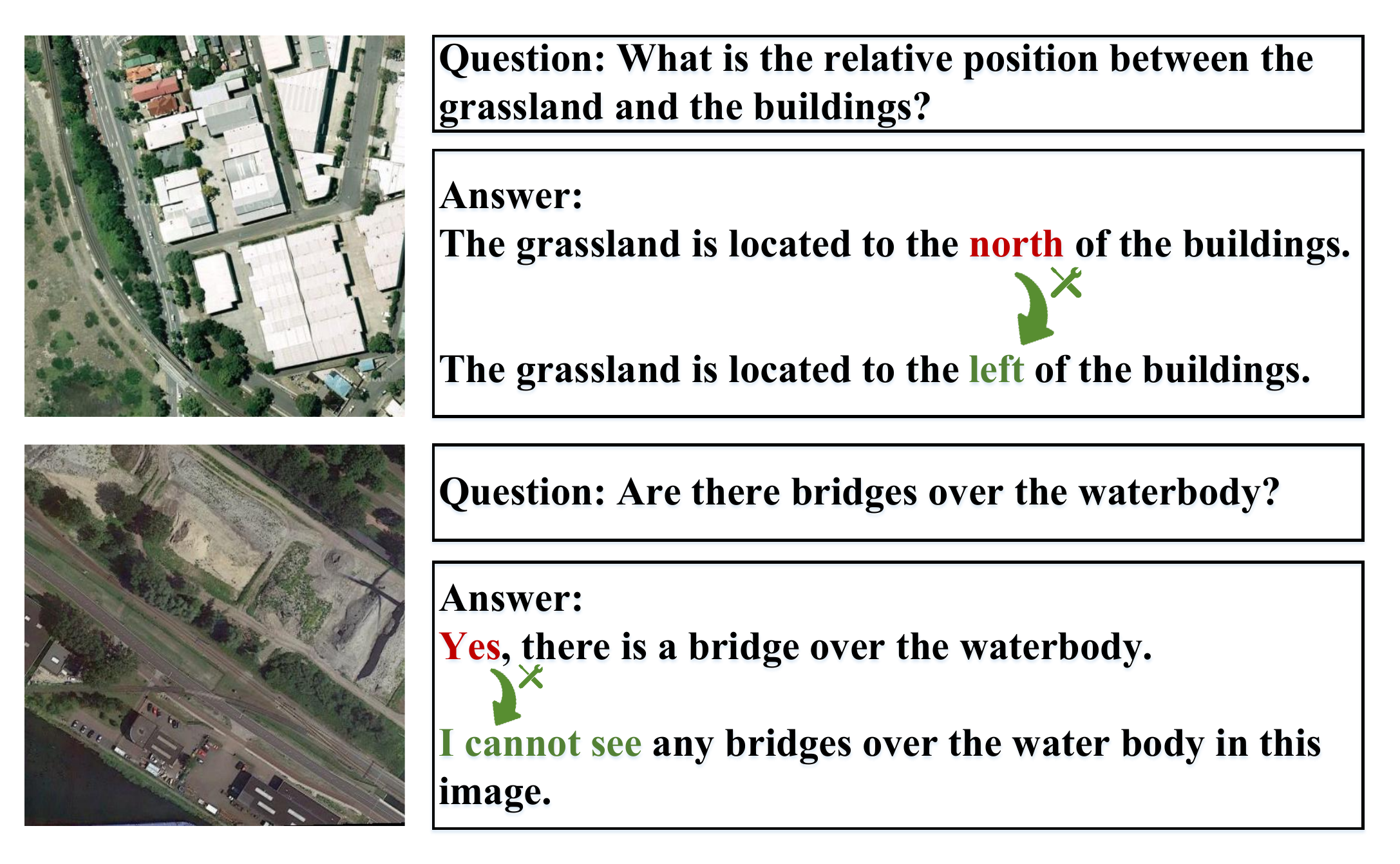}
\caption{Some examples of the performance of VHM on RSHalluEval before and after the decoding-based hallucination mitigation schemes. Red fonts indicate hallucinatory tokens, whereas green fonts indicate revised tokens.}
\label{visualize3}
\end{figure}

Additionally, Fig. \ref{visualize3} illustrates examples of applying the decoding-based strategy on VHM to alleviate hallucination. In the upper example, VHM correctly understands the spatial relationship between the grassland and buildings after the strategy is applied.

Moreover, the lower example demonstrates that the proposed strategy can alleviate snowballing hallucinations. Snowballing hallucination is a phenomenon where subsequent text generation is influenced by hallucinations in prior text, leading to an accumulation of hallucinations. In the example, when the strategy is absent, the first output token \enquote{Yes} is hallucinatory, causing subsequent tokens to elaborate on the number and location of the bridge based on this false premise. With the strategy applied, the first token is revised to \enquote{I}, enabling subsequent tokens to correctly judge the existence of bridges. These examples validate the effectiveness of the proposed strategy.

\section{Conclusion}
\label{sec:VIII}
This paper presents \textbf{RSHallu}, a systematic study of hallucinations in remote-sensing MLLMs.
We formalize RS hallucinations with an RS-oriented taxonomy and introduce \emph{image-level hallucination} to capture RS-specific inconsistencies beyond object-centric errors.
Building on this formalization, we develop a hallucination benchmark \textbf{RSHalluEval} together with a \emph{dual-mode} evaluation protocol that supports high-precision cloud auditing and low-cost, reproducible local checking.
To mitigate hallucinations, we construct \textbf{RSHalluShield} for training-friendly fine-tuning and propose training-free plug-and-play strategies, including decoding-time logit correction and RS-aware prompting.
Extensive experiments demonstrate consistent improvements in hallucination-free performance while maintaining competitive accuracy on downstream RS tasks, supporting more trustworthy deployment of RS-MLLMs in practical applications.

\section*{CRediT authorship contribution statement}
\textbf{Zihui Zhou:} Writing – original draft, Visualization, Validation, Methodology, Formal analysis, Conceptualization. \textbf{Yong Feng:} Writing – review \& editing, Supervision, Project administration, Funding acquisition, Conceptualization. \textbf{Yanying Chen:} Validation, Investigation. \textbf{Guofan Duan:} Investigation, Data curation. \textbf{Zhenxi Song:} Writing – review \& editing, Supervision. \textbf{Mingliang Zhou:} Validation, Conceptualization. \textbf{Weijia Jia:} Methodology, Conceptualization. 

\section*{Declaration of Competing Interest}
The authors declare that they have no known competing financial interests or personal relationships that could have appeared to influence the work reported in this paper.

\section*{Acknowledgements}
This work is supported by the National Nature Science Foundation of China under Grant No. 62262006, the Guangxi Science and Technology Key Research and Development Program under Grant No. AB24010112, the Heavy Rainfall Research Foundation of China under Grant No. BYKJ2024Z12, and the Key Project of Science and Technology Research Program of Chongqing Education Commission of China under Grant No. KJZD-K202402501.


\bibliographystyle{unsrt}
\bibliography{reference}




\end{document}